\pdfoutput=1

\documentclass[11pt]{article}
\usepackage[dvipsnames]{xcolor}
\usepackage{acl}

\usepackage{times}
\usepackage{latexsym}

\usepackage[T1]{fontenc}

\usepackage[utf8]{inputenc}

\usepackage{microtype}

\usepackage{inconsolata}
\usepackage{url}

\usepackage[most]{tcolorbox}

\colorlet{lightSalmon}{Salmon!80}

\usepackage{capt-of}
\usepackage{graphicx}
\usepackage{subfigure}
\usepackage{wrapfig}
\usepackage{tcolorbox}
\tcbuselibrary{listings, breakable}
\usepackage{amsmath}
\usepackage{amstext}
\usepackage{amsfonts}
\usepackage{bm}

\usepackage{bbm}
\definecolor{TrueLightPurple}{HTML}{E6E6FA} 
\usepackage{multirow}
\usepackage{booktabs}
\usepackage{array}
\usepackage{caption}
\usepackage{color}
\usepackage{colortbl}
\usepackage{tablefootnote}
\usepackage{adjustbox}
\usepackage{placeins}
\usepackage{tabularx}

\usepackage{multicol}
\usepackage{makecell} 
\usepackage{nicematrix}

\usepackage{algorithm}
\usepackage{algpseudocode}

\algnewcommand{\LineComment}[1]{\Statex ~~~~~~\textsc{//}~\textit{#1}}

\usepackage{enumitem}
\setenumerate[1]{itemsep=0pt,partopsep=0pt,parsep=\parskip,topsep=5pt}
\setitemize[1]{itemsep=0pt,partopsep=0pt,parsep=\parskip,topsep=5pt}
\setdescription{itemsep=0pt,partopsep=0pt,parsep=\parskip,topsep=5pt}

\usepackage{soul}

\usepackage{tikz}
\usepackage[edges]{forest}
\definecolor{hidden-draw}{RGB}{64,101,149}
\definecolor{hidden-pink}{RGB}{231,239,250}

\usepackage[mathscr]{euscript}

\usepackage{amssymb}  
\usepackage{pifont}

\definecolor{ForestGreen}{rgb}{0.13, 0.55, 0.13}
\definecolor{WildStrawberry}{rgb}{1.0, 0.26, 0.64}

\usepackage{xspace}

\newcommand{\method}{UniToolCall\xspace}

\usepackage[normalem]{ulem}

\title{
    \raisebox{-0.3\height}{\includegraphics[height=2.2em]{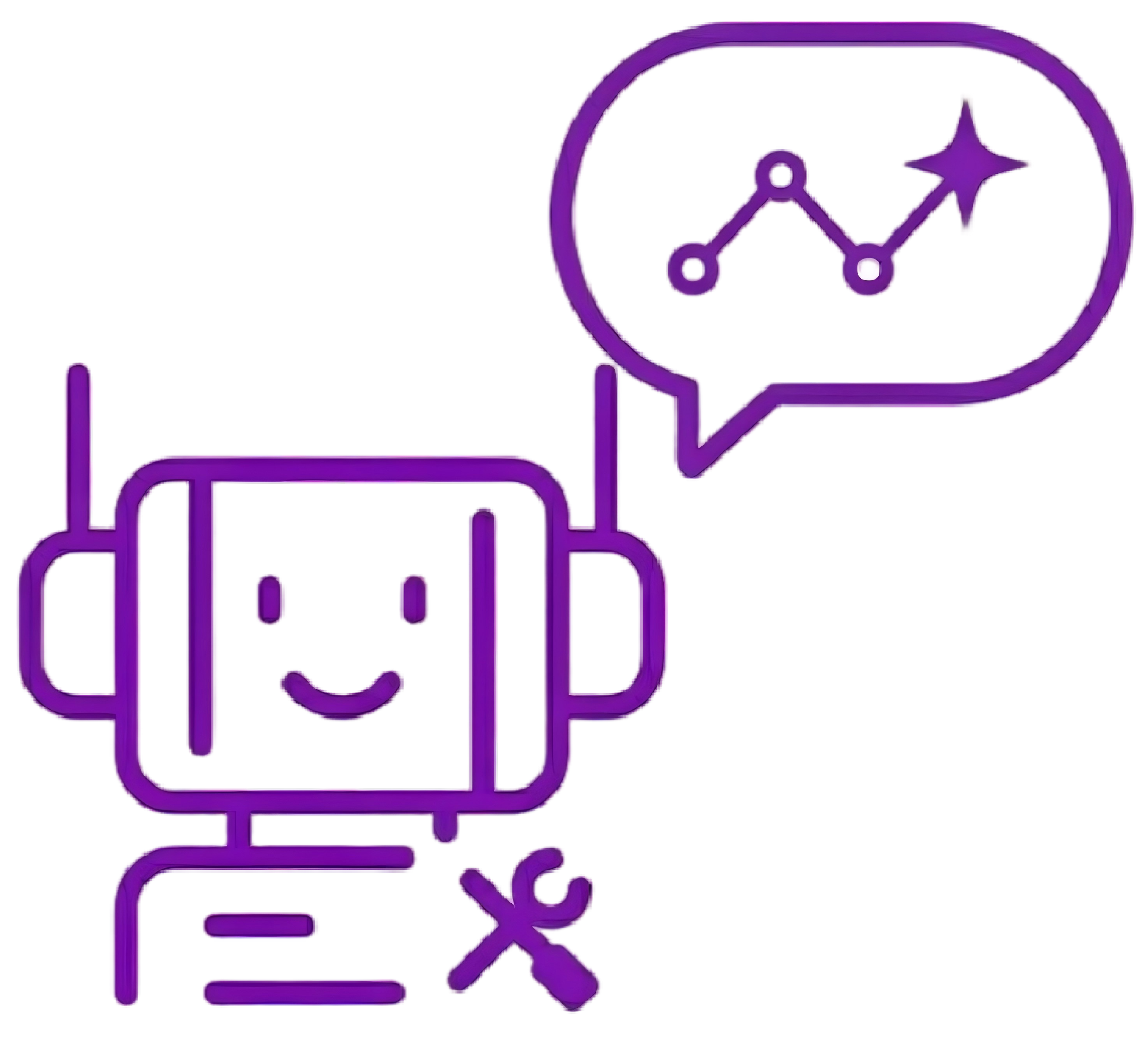}}UniToolCall: Unifying Tool-Use Representation, Data, and Evaluation for LLM Agents
}

\author{{Yijuan Liang}\textsuperscript{\rm 1,2}, {\textbf{Xinghao Chen}}\textsuperscript{\rm 2,3}, {\textbf{Yifan Ge}}\textsuperscript{\rm 2}, {\textbf{Ziyi Wu}}\textsuperscript{\rm 2}, {\textbf{Hao Wu}}\textsuperscript{\rm 2}, {\textbf{Changyu Zeng}}\textsuperscript{\rm 2}\\{\textbf{Wei Xing}}\textsuperscript{\rm 2},  {\textbf{Xiaoyu Shen}}\textsuperscript{\rm 2}\thanks{Corresponding Author}\\
  \textsuperscript{\rm 1} University of Science and Technology of China \\
  \textsuperscript{\rm 2} Zhejiang Key Laboratory of Industrial Intelligence and Digital Twin,\\
  \phantom{\textsuperscript{\rm 2} }Eastern Institute of Technology, Ningbo \\
  \textsuperscript{\rm 3} Department of Computing, The Hong Kong Polytechnic University \\
  \href{mailto:xyshen@eitech.edu.cn}{\texttt{xyshen@eitech.edu.cn}}
}

\begin{document}
\AtBeginEnvironment{thebibliography}{\hypersetup{hidelinks}}
\AtEndEnvironment{thebibliography}{\hypersetup{
    hidelinks=false,
    linkcolor=NavyBlue,
    citecolor=NavyBlue,
    urlcolor=NavyBlue
}}
\maketitle

\begin{abstract}
Tool-use capability is a fundamental component of LLM agents, enabling them to interact with external systems through structured function calls. However, existing research represents the query-to-answer chain inconsistently, rarely studies serial/parallel and hop/turn effects, and evaluates under incompatible protocols and metrics. 
We present \method, a unified framework for tool learning that standardizes the entire pipeline from toolset construction and dataset generation to evaluation. The framework curates a large tool pool of \textbf{22k+} tools and constructs a hybrid training corpus of \textbf{390k+} instances by combining 10 standardized public datasets with structurally controlled synthetic trajectories. It explicitly models diverse interaction patterns, including single-hop vs. multi-hop and single-turn vs. multi-turn, while capturing both serial and parallel execution structures. To support coherent multi-turn reasoning, we further introduce an Anchor Linkage mechanism that enforces cross-turn dependencies. Furthermore, we convert 7 public benchmarks into a unified \emph{Query--Action--Observation--Answer (QAOA)} representation with fine-grained evaluation at the function-call, turn, and conversation levels.
Experiments show that fine-tuning Qwen3-8B on our dataset substantially improves tool-use performance. Under the distractor-heavy Hybrid-20 setting, \method achieves \textbf{93.0\%} single-turn Strict Precision, outperforming commercial models including GPT, Gemini, and Claude. 
\footnote{\url{https://github.com/EIT-NLP/UniToolCall}.}

\end{abstract}

\section{Introduction}
\label{sec:introduction}

\begin{figure}[t]
    \centering
    \includegraphics[width=\columnwidth]{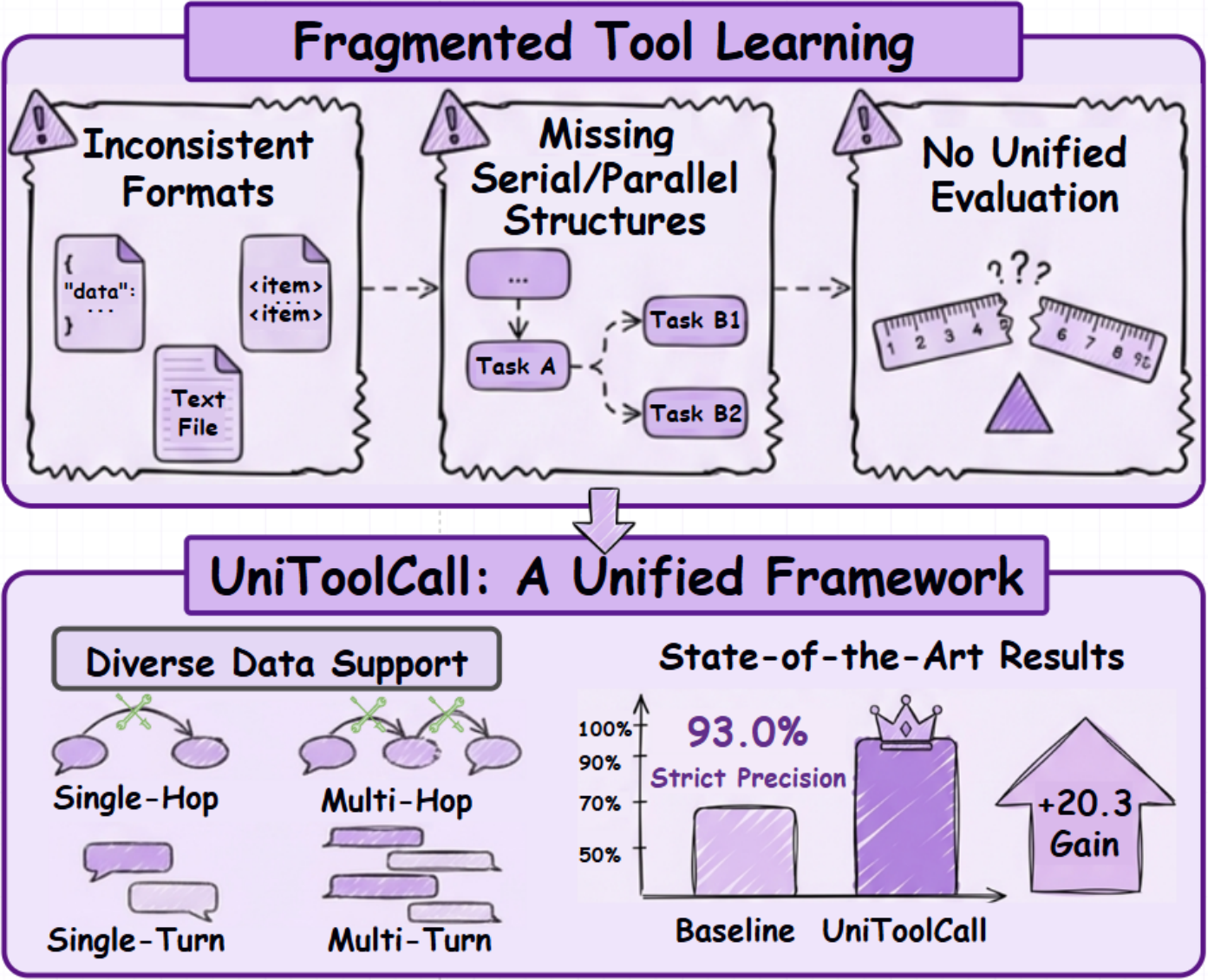} 
    \caption{Existing datasets are severely limited by the fragmentation problem. To address these challenges, \method introduces a standardized framework that provides robust structural constraints, yielding SOTA over strong baselines.}
    \label{fig:teaser}
\end{figure}
The emergence of LLM agents marks a shift from passive text generation to goal-directed interaction with external environments~\citep{durante2024agentaisurveyinghorizons,luo2025large,Sapkota_2026}. A key capability underlying this shift is tool use, which enables agents to take actions by translating natural language instructions into executable function calls. Through tool use, LLM agents can access external knowledge, invoke APIs, and perform multi-step operations, extending their capabilities beyond parametric knowledge~\citep{schick2023toolformerlanguagemodelsteach,yu2025survey}. Consequently, an agent’s effectiveness largely depends on its ability to select, compose, and execute tools reliably, making tool learning a central problem in agent research~\citep{paprunia2025advancing,lu2026tools}.

In the current data-driven paradigm, progress in tool learning is largely determined by the availability and quality of training data, particularly tool-use trajectories that capture how agents interact with external environments. Early efforts such as ToolLLM~\citep{qin2023toolllm}, ToolBench~\citep{patil2024gorilla}, and API-Bank~\citep{li2023apibank} construct such data by executing real-world APIs. While providing realistic supervision signals, they suffer from limited scalability and instability due to their reliance on external systems. To address these limitations, more recent works have shifted toward synthetic data generation, building simulated tool environments and automatically generating interaction trajectories (e.g., ToolForge~\citep{chen2025toolforge}, LoopTool~\citep{zhang2025looptool}, ASTRA~\citep{tian2026astraautomatedsynthesisagentic}). In parallel, a number of benchmarks have been proposed to evaluate tool-use capability, including ComplexFuncBench~\citep{zhong2025complexfuncbench}, HammerBench~\citep{wang2025hammerbench}, and ACEBench~\citep{chen2025acebench}.

Despite this progress, existing efforts are largely developed in isolation, leading to a fundamental fragmentation problem in tool learning. This fragmentation manifests along three key dimensions.
First, \emph{representation inconsistency}: different datasets adopt incompatible schemas to encode tool calls, arguments, and observations, making joint training across sources difficult.
Second, \emph{structural incompleteness}: current pipelines largely overlook the diversity of execution structures, particularly the distinction between serial and parallel tool invocation patterns.
Third, \emph{evaluation mismatch}: existing benchmarks rely on disparate protocols, tool definitions, and evaluation scripts, preventing fair and reproducible cross-dataset comparisons.
Together, these issues hinder both scalable training and systematic evaluation of tool-use capabilities.

To address these limitations, we propose \textbf{\method}, a unified framework for tool learning that standardizes the entire pipeline from toolset construction and data generation to evaluation under a shared representation. We first curate a large-scale tool pool by aggregating tools from multiple sources, resulting in a filtered set of over 22K tools. Building on this, we construct a hybrid training corpus that combines standardized public datasets with structurally controlled synthetic trajectories, yielding 390K instances spanning single-hop, multi-hop, single-turn, and multi-turn interactions. Crucially, our synthetic pipeline explicitly models both serial and parallel execution structures, enabling fine-grained analysis of execution patterns.
Finally, we unify all data into a Query–Action–Observation–Answer (QAOA) representation~\footnote{A single-hop sample is shown in Appendix~\ref{sec:data_sample}.} and introduce a standardized evaluation protocol with comprehensive metrics, enabling consistent and fair comparison across diverse settings. Our contributions are as follows:

\begin{itemize}
\item \textbf{Structurally-Aware data generation} 
We propose a synthetic data generation pipeline that provides controlled supervision for single/multi-hop and single/multi-turn interactions. The pipeline explicitly models both serial and parallel execution patterns and introduces an \emph{Anchor Linkage} mechanism to enforce cross-turn dependencies.

\item \textbf{A standardized unified benchmark} 
We convert heterogeneous public benchmarks into a unified QAOA format with shared matching rules and metrics, enabling fine-grained evaluation across function-call, turn, and conversation levels and fair comparison across diverse task structures.

\item \textbf{Strong empirical performance} Fine-tuning Qwen3-8B on our framework achieves state-of-the-art results. Under the distractor-heavy Hybrid-20 setting, \method attains \emph{93.0\%} single-turn Strict Precision, outperforming leading commercial models including GPT, Gemini, and Claude.
\end{itemize}

\section{Related work}
\label{sec:related_work}

\paragraph{Synthetic data generation}
Early work~\citep{tang2023toolalpaca} generates instruction-style tool-use examples to teach basic API usage. Subsequent pipelines further automate dataset construction~\citep{liu2024apigen, chen2025toolforge, zhang2025looptool}, synthesizing tool-use trajectories at a larger scale. Recently, advanced pipelines such as ToolACE~\citep{liu2025toolace}, ToolMind~\citep{yang2025toolmind}, APIGen-MT~\citep{prabhakar2026apigenmt}, and BUTTON~\citep{chen2025facilitating} utilize multi-agent simulated interplay to generate complex dialogues, while MAGNET~\citep{magnet}, ToolDial~\citep{shim2025tooldial}, and ToolFlow~\citep{toolflow} rely on graph-based routing for multi-hop synthesis. Despite this progress, generated trajectories still exhibit structural blind spots: they lack explicit control over serial versus parallel execution, and rely heavily on the LLM's inherent stability for multi-turn state tracking, risking context drift. In contrast, our pipeline strictly parameterizes structural diversity across four interaction structures. Furthermore, we introduce an Anchor Linkage mechanism with deterministic fallbacks to guarantee strict cross-turn state inheritance. A structured comparison with representative synthesis pipelines and TRAJECT-Bench is provided in Appendix~\ref{sec:related_resources}.

\paragraph{Tool-use benchmarks}
Some studies rely on real environments~\citep{qin2023toolllm, wang2025mcpbench, gao2025mcpradar}, where models interact with external tools through actual execution. To improve reproducibility, several benchmarks evaluate tool usage through simulated invocation while retaining real tool definitions~\citep{chen2025acebench, moon2024toolbank}. Concurrently, TRAJECT-Bench~\citep{he2025traject} advances structure-aware evaluation by modeling parallel and sequential trajectories with controlled scaling and trajectory-level diagnostics beyond final-answer accuracy. However, these benchmarks adopt heterogeneous schemas, evaluation rules, and task structures, which hinder fair comparison. To address these limitations, we construct a unified benchmark that evaluates heterogeneous datasets under a shared QAOA representation, enabling multi-granularity evaluation and providing a more comprehensive assessment of tool learning.
\begin{figure*}[t]
    \centering
    \includegraphics[width=\textwidth]{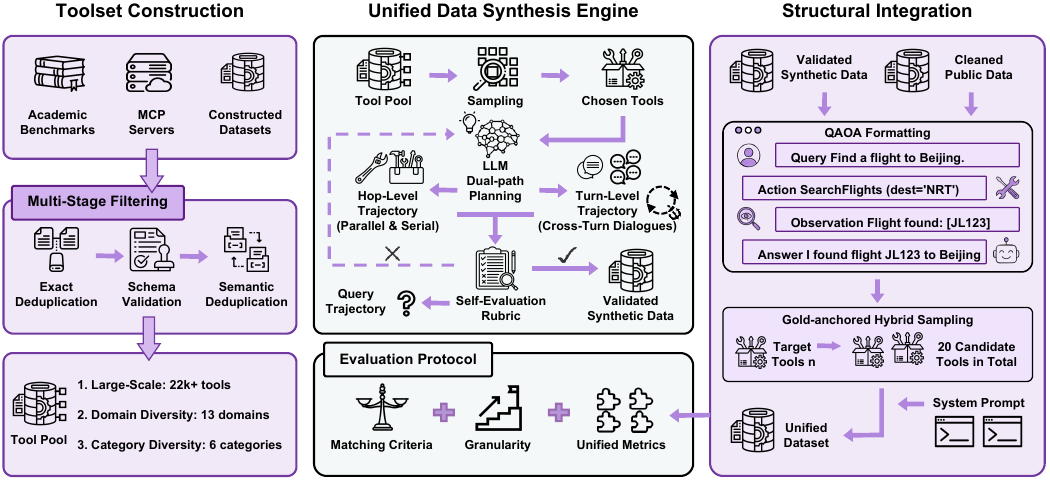} 
    \caption{The overall architecture of \method, comprising several interconnected modules: (1) Toolset construction; (2) Unified data synthesis engine; (3) Structural integration; (4) Evaluation protocol.}
    \label{fig:main_pipeline}
\end{figure*}

\section{\method}

In this section, we present \method, a unified framework for tool learning. At the core of our framework is a standardized QAOA representation, which provides a consistent format for modeling tool interactions across datasets. As illustrated in Figure~\ref{fig:main_pipeline}, the framework consists of three components: a curated toolset, a data generation pipeline, and a structural assembly stage. In addition, we introduce a unified benchmark to enable consistent evaluation across tool-use scenarios.

\subsection{Toolset construction}
\label{sec:toolset_construction}

To serve as the candidate pool for dataset construction, we construct a comprehensive toolset after applying the multi-stage filtering mechanism (Appendix~\ref{sec:appendix_tool_filtering}), denoted as $\mathcal{T}$. As illustrated in Figure~\ref{fig:toolset}, the toolset is formed from three primary sources: (1) Academic benchmarks, (2) MCP servers, and (3) Constructed datasets. All tools are standardized into a unified JSON Schema format. To facilitate semantic organization and balanced sampling during data generation, we categorize tools along two dimensions: functional \textbf{category} and application \textbf{domain}. Based on common API usage patterns in agent systems, we define 6 functional categories (e.g., visualization, analysis) to capture the operational roles of tools and 13 application domains (e.g., finance, technology) to represent typical real-world usage scenarios.\footnote{The details, complete taxonomy and category definitions are provided in Appendix~\ref{sec:toolset} and~\ref{sec:appendix_taxonomy}.}

\subsection{Training dataset construction}
\label{sec:train_construction}

\paragraph{Source} To equip the agent with robust tool-use and planning capabilities, we construct a large-scale hybrid training dataset, denoted as $D_{\text{train}}$. The dataset is composed of two parts: (1) \textbf{Public data integration ($D_{\text{pub}}$)} We collect and integrate 10 distinct tool-use datasets\footnote{Table~\ref{tab:public_datasets} provides detailed statistics for each dataset.}. To ensure validity and reliability, we implement a two-stage filtering strategy applied before and after format conversion (Appendix~\ref{sec:public_summary}). (2) \textbf{Synthetic augmentation ($D_{\text{syn}}$)} To overcome the structural shallowness inherent in public corpora, we construct a synthetic dataset $D_{\text{syn}}$ based on the toolset $\mathcal{T}$. In particular, the pipeline controls both execution patterns (serial vs. parallel tool invocation) and interaction complexities (single-hop, multi-hop, single-turn, and multi-turn scenarios). $D_{\text{syn}}$ is filtered using an LLM-based self-evaluation framework using six core metrics (e.g., Tool-fit, Success), plus an additional anchor-linkage metric for multi-turn episodes (Appendix~\ref{sec:synthetic_eval}).

\begin{figure}[htbp]
    \centering
    \includegraphics[width=\columnwidth]{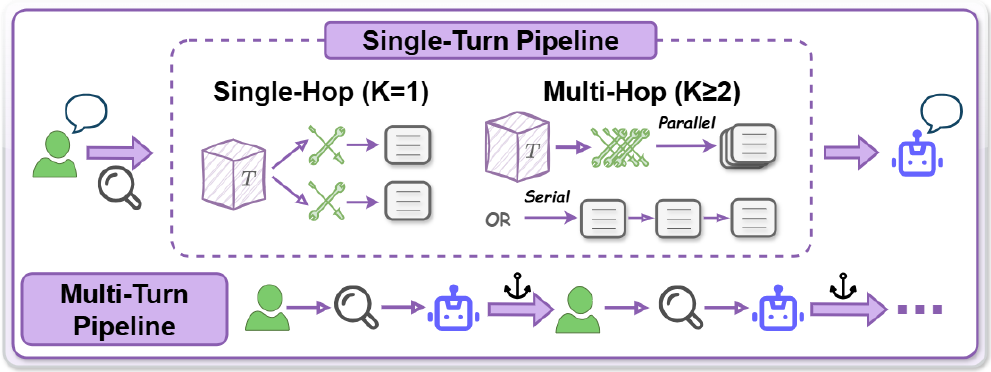} 
    \caption{Synthetic trajectory generation: single-hop and serial/parallel multi-hop in the single-turn pipeline, and multi-turn generation with Anchor Linkage.}
    \label{fig:pipeline}
\end{figure}

\paragraph{Unified synthetic data pipeline} We design a unified generative framework equipped with stringent quality control (Figure~\ref{fig:pipeline}). Formally, the construction of any synthetic subset $D_{x} \in \{D_{\text{sh}}, D_{\text{mh}}, D_{\text{mt}}\}$ is generalized as follows:
{\setlength{\abovedisplayskip}{6pt plus 2pt minus 2pt}%
\setlength{\belowdisplayskip}{6pt plus 2pt minus 2pt}%
\setlength{\abovedisplayshortskip}{0pt}%
\setlength{\belowdisplayshortskip}{4pt plus 1pt minus 1pt}%
\begin{equation*}
\resizebox{0.95\columnwidth}{!}{$\displaystyle
D_{x} = \big\{ \Psi (\tau, P_{\text{sys}}, L_{\text{cand}}) \mid S \subseteq \mathcal{T},\; \tau \sim \mathcal{M}(S),\; \Phi_{\text{eval}}(\tau) = 1 \big\}
$}
\end{equation*}}
where $S$ is a sampled subset from the filtered tool pool $\mathcal{T}$, $\tau$ represents the raw interaction trajectory generated by the LLM $\mathcal{M}$, and $\Phi_{\text{eval}}$ acts as the heuristic self-evaluation gate. Across all scenarios, the structural assembly function $\Psi$ standardizes the validated trajectories into our QAOA format. Crucially, $\Psi$ constructs the candidate list $L_{\text{cand}}$ using a uniform \textbf{Hybrid-20 setting}: retaining the ground-truth tools from $S$ as anchors, retrieving top-ranking hard negatives via embedding similarity, and appending 5 random easy negatives to yield exactly 20 candidates~\citep{themcpcompany2025}. Finally, a system prompt $P_{\text{sys}}$ (Appendix~\ref{sec:data_sample}) detailing tool-use constraints is injected. While sharing this core formulation, the specific definitions of the tool subset $S$ and the trajectory $\tau$ diverge to target distinct agentic capabilities:

\paragraph{Single-Hop ($D_{\text{sh}}$)} 
For fundamental invocation mapping, we sample a single tool ($|S| = 1$) and let $\mathcal{M}$ generate a one-step trajectory $\tau = \langle q, a, o, r \rangle$ from the tool schema, where $q$, $a$, $o$, and $r$ denote Query, Action, Observation, and Answer.

\paragraph{Multi-Hop ($D_{\text{mh}}$)} 
To train coordination over sequences of tool calls, we sample a domain-constrained subset $S$ ($|S| \in \{2,\dots,5\}$) and generate a $K$-step trajectory ($K \ge 2$) with explicit control over execution routing. For \textbf{serial} instances, $\mathcal{M}$ generates steps iteratively, constraining subsequent hops to reference concrete values from earlier observations to form genuine inter-step dependencies. For \textbf{parallel} instances, the query $q$ and all tool calls are synchronized in a one-shot generation to prevent intention-tool mismatches.

\paragraph{Multi-Turn ($D_{\text{mt}}$)} 
For long-horizon, stateful interactions across $T \in \{2,3,4\}$ dialogue turns, we sample a usage-balanced subset $S$ ($|S| = 10$). Generation requires a \textbf{Two-Stage planning} mechanism (episode-level storyline followed by turn-level intent). Furthermore, to address the disjointed context shifts common in existing multi-turn corpora \citep{ma2024agentboard}, we introduce \textbf{Anchor Linkage}, a hard constraint that requires each user query at turn $t \ge 2$ to explicitly inherit concrete state variables (e.g., transaction IDs, coordinates, or returned parameter values) from turn $t-1$. Without such constraints, unconstrained generators often produce locally fluent but context-disjoint turns, a failure mode that generic fluency metrics fail to penalize (Section~\ref{sec:ablations}; Appendix~\ref{sec:appendix_anchor_failures}). Anchor Linkage combines three stages: (1)~\textbf{dynamic prompting} that escalates from soft anchor hints to hard inclusion constraints on retries; (2)~\textbf{programmatic verification} with regex-based checks and a string-prepending fallback; and (3)~\textbf{evaluation gating} via the $s_{\text{anchor}}$ rubric so inherited anchors remain functionally meaningful, not merely copied. \footnote{Implementation details and qualitative failure cases without the constraint are in Appendices~\ref{sec:appendix_anchor_linkage} and~\ref{sec:appendix_anchor_failures}.}

\subsection{Evaluation protocol}
\label{sec:test_dataset}

To evaluate model performance across complex tool-use scenarios, we construct a unified benchmark $D_{\text{test}}$ focusing on two fundamental capabilities: accurate tool selection and correct parameter generation. Converting all raw data into the standardized QAOA framework enables fairer comparisons across datasets.\footnote{Field-level preserve/normalize decisions and per-benchmark sample retention are detailed in Appendix~\ref{sec:benchmark_conversion}.} Since agent interactions exhibit a hierarchical structure, where conversations consist of multiple turns, each containing function calls, we decouple our evaluation into the following granularities:

\paragraph{Function call-level verification}
At the most fundamental level, the validity of each individual tool invocation is assessed by matching the predicted tool name and generated arguments against the ground truth. This call-level correctness serves as the computational basis for calculating proportional scores in flexible metrics.

\paragraph{Turn \& Conversation-level aggregation}
To evaluate task-level capabilities, the aforementioned call-level results are aggregated at higher dimensions, denoted by $N$. We explicitly map the aggregation granularity to the specific type of task complexity being assessed: (1) \textbf{Turn-level:} For single/multi-hop scenarios, we compute metrics across individual dialogue turns. (2) \textbf{Conversation-level:} For single/multi-turn scenarios, we compute metrics across the entire dialogue trajectory. During aggregation, Strict metrics employ an all-or-nothing penalty (the instance scores 0 if any function call is flawed), whereas Flexible metrics award credit based on the ratio of correct function calls within the instance. These instance scores are subsequently macro-averaged across the dataset.

\paragraph{Matching criteria}

At the Function Call-level, we employ a cascaded strategy to determine the validity of a predicted function call: (1) \textbf{Rule-based matching:} Serving as the primary strategy, this method achieves exact matching through rigorous standardization\footnote{Refer to Appendix~\ref{sec:appendix_matching_rules} for the detailed rules.}. A match is confirmed if the standardized prediction aligns perfectly with the ground truth. (2) \textbf{Semantic matching:} We calculate the ROUGE-L similarity score between the prediction and the reference. A prediction is deemed a semantic match if the score is $\ge 0.7$.

\label{sec:metric_definitions}
\section{Experiments}

\subsection{Experimental setup}
\label{sec:exp_setup}

\paragraph{Models and training data}
We instantiate \method on the open-source backbone Qwen3-8B~\cite{yang2025qwen3}. The model is fine-tuned on our comprehensive dataset $D_{\text{train}}$, which contains 390,060 instances grounded in our tool pool $\mathcal{T}$ of 22,606 tools. Specifically, this consists of 387,123 high-quality conversations retained from the standardized public corpus ($D_{\text{pub}}$), and 2,937 synthetic trajectories ($D_{\text{syn}}$). The synthetic subset is generated and rigorously self-evaluated by Qwen3-32B~\cite{yang2025qwen3}, comprising 979 instances for each of the single-hop, multi-hop, and multi-turn scenarios respectively. 

\paragraph{Implementation details}
We employ the LLaMA-Factory framework integrated with DeepSpeed optimization to fine-tune the base model. Training is conducted with LoRA~\cite{hu2022lora}. We target all linear modules with rank $r=8$ and scaling factor $\alpha=16$. The model is trained for 1 epoch using AdamW with a learning rate of $1\times10^{-5}$ and a warmup ratio of $0.03$. The maximum sequence length is set to 8192 tokens. The effective batch size is 8. We use bfloat16 precision throughout training. All experiments are conducted on a single node with 4$\times$NVIDIA A800-SXM4 GPUs (40GB each) and an Intel Xeon Platinum 8378A CPU.

\paragraph{Baselines}
We compare \method against six strong LLM baselines: GPT-5.2 Instant~\cite{openai2025gpt52}, Gemini 3 Flash Preview~\cite{google2025gemini3}, Claude 4.6 Sonnet~\cite{anthropic2025claude46}, Kimi-K2-Instruct~\cite{kimiteam2025k2}, DeepSeek-V3.2~\cite{deepseek2025v32}, and Qwen3-32B~\cite{yang2025qwen3}. To improve inference efficiency, we standardize the inference setup by disabling explicit reasoning traces when supported (e.g., \texttt{<think>} style outputs). Our fine-tuned Qwen3-8B is trained and evaluated with \texttt{enable-thinking=false}, and DeepSeek-V3.2, Qwen3-32B, and Claude 4.6 Sonnet are evaluated with reasoning disabled through the available API options. For the remaining proprietary models, we use their non-reasoning or efficiency-oriented variants. 

\paragraph{Evaluation settings}
Our main evaluation is conducted under the Hybrid-20 setting mentioned in Section~\ref{sec:train_construction}. We additionally report results under the Ground Truth (GT) setting, in which the candidate list contains only the required target tools. The main result tables report a single representative run for each model. For Qwen3-8B Vanilla and \method, Table~\ref{tab:repeated_run_stability} reports three independent training runs (mean $\pm$ std): the gains hold across runs, and \method exhibits smaller variance on the main single/multi-hop and single-turn metrics. 

The unified evaluation $D_{\text{test}}$ mentioned in Section~\ref{sec:test_dataset} comprises 7 public benchmarks, yielding 6,163 high-quality conversations after filtering (Table~\ref{tab:public_datasets}). Based on this evaluation protocol, we define four core quantitative metrics. We first introduce three indicator functions for a predicted function call $p$ within an instance's prediction set $P_i$ (Missed cases will be filled with null), evaluated against the ground truth set $G_i$:
\begin{itemize}
    \item $m_{n}(p)$: Returns 1 if $p$ has a correctly matching tool name in $G_i$; otherwise 0.
    \item $m_{s}(p)$: Returns 1 if $p$ strictly matches a call in $G_i$ in both name and all argument values.
    \item $m_{f}(p)$: Returns 1 if $p$ matches the tool name and satisfies the semantic similarity threshold for arguments.
\end{itemize}
\textit{Note: Under-predicted calls in $P_i$ are padded with null to penalize omissions, directly yielding a score of $0$ when no tools are invoked ($|P_i|=0$).}

\paragraph{Strict Precision (SP)} 
This metric establishes the rigorous lower bound for tool selection. An instance scores 1 if and only if every predicted tool name perfectly matches the ground truth:
\begin{equation*}
    \scalebox{0.88}{$
        \text{SP} = \frac{1}{N} \sum_{i=1}^{N} \mathbf{1} \big( |P_i| = |G_i| \land \forall p \in P_i, m_{n}(p)=1 \big)
    $}
\end{equation*}

\paragraph{Flexible Precision (FP)} 
As a tolerant tool selection metric, this macro-averaged precision calculates the proportion of correctly named tools:
\begin{equation*}
    \scalebox{0.88}{$
        \text{FP} = \frac{1}{N} \sum_{i=1}^{N} \frac{1}{|P_i|} \sum_{p \in P_i} m_{n}(p)
    $}
\end{equation*}

\paragraph{Strict Parameter Accuracy (SPA)} 
This metric assesses the exactness of argument generation. The denominator is the total number of predicted calls ($|P_i|$). A prediction only contributes to the score if both its name and arguments are perfectly correct:
\begin{equation*}
    \scalebox{0.88}{$
        \text{SPA} = \frac{1}{N} \sum_{i=1}^{N} \frac{1}{|P_i|} \sum_{p \in P_i} m_{s}(p)
    $}
\end{equation*}

\paragraph{Flexible Parameter Accuracy (FPA)} 
This metric measures the proportion of predicted tools that pass argument matching: either exact rule-based matching or ROUGE-L similarity:
\begin{equation*}
    \scalebox{0.88}{$
        \text{FPA} = \frac{1}{N} \sum_{i=1}^{N} \frac{1}{|P_i|} \sum_{p \in P_i} m_{f}(p)
    $}
\end{equation*}

For our primary evaluation, we report micro-average to reflect the natural volume distribution of the collected public benchmarks. However, recognizing the inherent sample size imbalance across different source datasets, we additionally conduct a dataset-level macro-average analysis in Appendix~\ref{sec:appendix_imbalance}. This supplementary evaluation grants equal weight to each sub-benchmark, ensuring our observed performance gains are robust.

\begin{table*}[t]
\centering
\small
\setlength{\tabcolsep}{3pt} 
\resizebox{\textwidth}{!}{%
\begin{tabular}{l c cccc c cccc c cccc c cccc} 
\toprule
\multicolumn{1}{c}{\multirow{2}{*}{\textbf{Models}}} & & \multicolumn{4}{c}{\textbf{SP (\%)~$\uparrow$}} & & \multicolumn{4}{c}{\textbf{FP (\%)~$\uparrow$}} & & \multicolumn{4}{c}{\textbf{SPA (\%)~$\uparrow$}} & & \multicolumn{4}{c}{\textbf{FPA (\%)~$\uparrow$}} \\
\cmidrule(lr){3-6} \cmidrule(lr){8-11} \cmidrule(lr){13-16} \cmidrule(lr){18-21}
 & & SH & MH & ST & MT & & SH & MH & ST & MT & & SH & MH & ST & MT & & SH & MH & ST & MT \\
\midrule
\rowcolor{gray!15} \multicolumn{21}{c}{\textit{\textbf{GT Setting}}} \\
\midrule
Qwen3-8B (Upper Bound) & & 96.1 & 47.5 & 92.6 & 0.0 & & 96.1 & 76.4 & 95.5 & 39.5 & & 28.7 & 57.2 & 32.1 & 18.8 & & 52.3 & 70.0 & 54.9 & 24.9 \\
\midrule
\rowcolor{gray!15} \multicolumn{21}{c}{\textit{\textbf{Hybrid-20 Setting}}} \\
\midrule
\textbf{Proprietary Models} & & & & & & & & & & & & & & & & & & & & \\
GPT-5.2 Instant & & 50.9 & 39.0 & 50.5 & 0.0 & & 50.9 & 58.2 & 52.5 & 16.1 & & 23.2 & 49.1 & 26.3 & 9.1 & & 39.2 & 55.3 & 41.5 & 9.8 \\
Gemini 3 Flash Preview & & 68.3 & 77.2 & 70.3 & 0.0 & & 68.3 & 83.6 & 70.9 & 22.5 & & \underline{25.2} & 69.3 & 30.3 & 14.2 & & \underline{46.3} & 78.5 & \underline{50.5} & 15.6 \\
Claude 4.6 Sonnet & & 58.8 & \textbf{83.3} & 62.1 & 0.0 & & 58.8 & \textbf{89.6} & 62.7 & 34.8 & & 24.9 & \textbf{74.1} & \underline{30.3} & 19.4 & & 43.1 & \textbf{84.1} & 47.9 & 25.2 \\
\midrule
\textbf{Open-Source Models} & & & & & & & & & & & & & & & & & & & & \\
Kimi-K2-Instruct & & 55.9 & 75.9 & 58.7 & 0.0 & & 55.9 & 86.1 & 59.7 & 31.1 & & 21.3 & \underline{69.6} & 26.5 & 18.4 & & 37.1 & \underline{80.1} & 42.0 & 24.9 \\
DeepSeek-V3.2 & & 46.9 & 39.7 & 47.0 & 0.0 & & 46.9 & 65.3 & 49.6 & 15.0 & & 18.8 & 54.3 & 22.9 & 10.3 & & 32.8 & 60.9 & 36.3 & 13.0 \\
Qwen3-32B & & \underline{72.6} & 64.0 & \underline{72.7} & \textbf{8.3} & & \underline{72.6} & 79.9 & \underline{74.3} & \underline{38.2} & & 23.7 & 61.3 & 27.9 & \textbf{21.6} & & 43.8 & 72.9 & 47.4 & \textbf{28.8} \\
Qwen3-8B (Vanilla) & & 66.9 & 22.7 & 63.3 & 0.0 & & 66.9 & 53.9 & 66.5 & 25.8 & & 19.8 & 38.1 & 22.1 & 12.1 & & 36.7 & 47.0 & 38.3 & 16.1 \\
\rowcolor{TrueLightPurple} 
\textbf{UniToolCall (Ours)} & & \makecell{\textbf{92.9} \\ \textcolor{ForestGreen}{\scriptsize $\uparrow$ 26.0}} & \makecell{\underline{80.7} \\ \textcolor{ForestGreen}{\scriptsize $\uparrow$ 58.0}} & \makecell{\textbf{93.0} \\ \textcolor{ForestGreen}{\scriptsize $\uparrow$ 29.7}} & \makecell{0.0 \\ \textcolor{WildStrawberry}{\scriptsize $\downarrow$ 0.0}} & & \makecell{\textbf{92.9} \\ \textcolor{ForestGreen}{\scriptsize $\uparrow$ 26.0}} & \makecell{\underline{89.6} \\ \textcolor{ForestGreen}{\scriptsize $\uparrow$ 35.7}} & \makecell{\textbf{93.8} \\ \textcolor{ForestGreen}{\scriptsize $\uparrow$ 27.3}} & \makecell{\textbf{39.4} \\ \textcolor{ForestGreen}{\scriptsize $\uparrow$ 13.6}} & & \makecell{\textbf{27.1} \\ \textcolor{ForestGreen}{\scriptsize $\uparrow$ 7.3}} & \makecell{66.6 \\ \textcolor{ForestGreen}{\scriptsize $\uparrow$ 28.5}} & \makecell{\textbf{31.6} \\ \textcolor{ForestGreen}{\scriptsize $\uparrow$ 9.5}} & \makecell{\underline{21.2} \\ \textcolor{ForestGreen}{\scriptsize $\uparrow$ 9.1}} & & \makecell{\textbf{48.6} \\ \textcolor{ForestGreen}{\scriptsize $\uparrow$ 11.9}} & \makecell{78.8 \\ \textcolor{ForestGreen}{\scriptsize $\uparrow$ 31.8}} & \makecell{\textbf{52.4} \\ \textcolor{ForestGreen}{\scriptsize $\uparrow$ 14.1}} & \makecell{\underline{26.1} \\ \textcolor{ForestGreen}{\scriptsize $\uparrow$ 10.0}} \\
\bottomrule
\end{tabular}%
}
\caption{Comprehensive evaluation results across varying tool-use complexities. SH, MH, ST, and MT denote Single-Hop, Multi-Hop, Single-Turn, and Multi-Turn scenarios, respectively. All reported metrics are scaled to percentages (\%). The best results are bolded and the second best results are underlined in all following tables.}
\label{tab:main_results}
\end{table*}

\begin{table}[ht]
\centering
\small
\setlength{\tabcolsep}{1.7pt} 
\resizebox{\columnwidth}{!}{%
\begin{tabular}{l c cc c}
\toprule
\multicolumn{1}{c}{\multirow{2}{*}{\textbf{Method}}} & \textbf{MH Ratio} & \multicolumn{2}{c}{\textbf{MH}} & \textbf{MT} \\
\cmidrule(lr){3-4} \cmidrule(lr){5-5}
 & \textbf{(Ser:Par)} & \textbf{SP (\%)$\uparrow$} & \textbf{SPA (\%)$\uparrow$} & \textbf{FP (\%)$\uparrow$} \\
\midrule
Vanilla Qwen3-8B & - & 22.7 & 38.1 & 25.8 \\
\midrule
\rowcolor{gray!15} \multicolumn{5}{c}{\textit{Ablation I: Public vs.\ Synthetic under matched budget}} \\
\midrule
Synthetic Mixed & 1 : 1.9 & \underline{57.1} & \underline{56.8} & \underline{39.6} \\
Public Mixed & 1 : 5.7 & \textbf{59.7} & \textbf{58.2} & \textbf{47.9} \\
\midrule
\rowcolor{gray!15} \multicolumn{5}{c}{\textit{Ablation II: Synthetic-only comparison (Homogeneous vs.\ Mixed)}} \\
\midrule
Pure Single-hop & - & 51.9 & 54.9 & 38.5 \\
Pure Multi-hop & 1 : 1.3 & 53.4 & 55.6 & \underline{42.1} \\
Pure Multi-turn & 1 : 0.9 & \underline{54.3} & \underline{56.4} & \textbf{44.4} \\
Synthetic Mixed & 1 : 1.9 & \textbf{57.1} & \textbf{56.8} & 39.6 \\
\bottomrule
\end{tabular}%
}
\caption{Controlled analysis under a matched data budget ($N=979$). MH Ratio denotes the proportion of serial to parallel multi-hop trajectories. Synthetic Mixed contains a mix of single-hop, multi-hop, and multi-turn data generated by our pipeline. Public Mixed is sampled exclusively from the public corpora $D_{\text{pub}}$ while preserving its original structural distribution. Pure denotes models trained entirely on one specific synthetic data type (e.g., Pure Single-hop).}
\label{tab:ablation_composition}
\end{table}
\begin{figure*}[t]
    \centering
    \includegraphics[width=\textwidth]{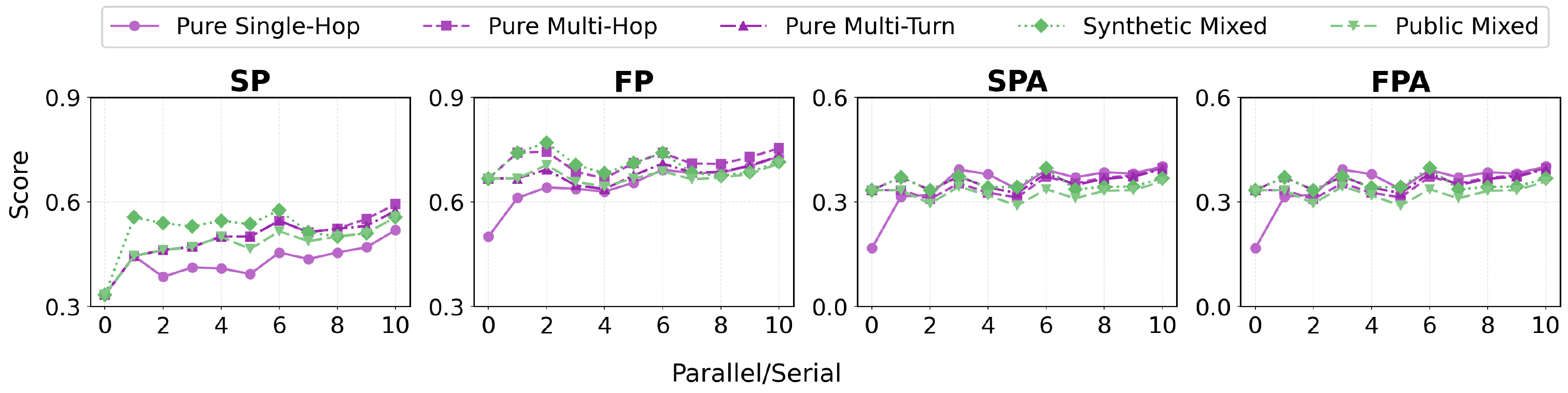} 
    \caption{Performance trends across different data compositions under varying parallel-to-serial ratios.}
    \label{fig:bfcl_ratio}
\end{figure*}

\subsection{Main results}
\label{sec:main-exp}

Table~\ref{tab:main_results} summarizes the overall performance of \method and the baselines under the Hybrid-20 setting. We highlight three main observations.

\paragraph{Strong gains in tool selection}
\method achieves the best strict tool-selection performance in both single-hop and single-turn settings, reaching SP scores of 92.9\% and 93.0\%, respectively. These results substantially improve over the vanilla Qwen3-8B and also exceed stronger open-source and proprietary baselines such as Qwen3-32B and Gemini 3 Flash Preview. In multi-hop settings, \method obtains the second-best SP (80.7\%) and FP (89.6\%), while remaining close to Claude 4.6 Sonnet on strict selection. This suggests that our framework is particularly effective at improving precise tool localization under distractor-heavy retrieval conditions.

\paragraph{Improved parameter grounding}
Beyond tool selection, \method also improves parameter generation quality. In the single-turn setting, it achieves the best SPA (31.6\%) and FPA (52.4\%) across all compared models. Multi-hop SPA and FPA likewise rise over vanilla (66.6\% / 78.8\% vs.\ 38.1\% / 47.0\%), while remaining behind Claude 4.6 Sonnet on those two cells. In multi-turn scenarios, \method improves FP and parameter-level metrics compared to the vanilla backbone, although strict conversation-level matching remains challenging for all models due to the long-horizon nature of the task.

\paragraph{Approaching GT upper bound}
The GT setting removes distractor tools and therefore serves as a useful reference point for analyzing retrieval difficulty. In single-hop evaluation, \method under Hybrid-20 approaches the vanilla model's GT performance. In single-turn and multi-hop settings, the fine-tuned model even surpasses the vanilla model evaluated in the GT condition, suggesting that the gains are not limited to distractor resistance alone, but also reflect improved intrinsic capability in structured tool-use prediction.

\paragraph{Long-horizon execution remains a challenge} 
Almost all models score 0.0\% on Multi-Turn SP under conversation-level strict aggregation: a single incorrect call zeros the entire episode. This reflects error accumulation across turns in distractor-heavy settings rather than an inability to solve any individual turn. Vanilla prefix survival on the BFCL multi-turn subset falls to 3.3\% by Turn 2, while \method remains at 50.0\% (Appendix~\ref{sec:appendix_mt_cliff}). Long-horizon strict success nonetheless remains an open challenge; FP is therefore essential for measuring partial progress, under which \method achieves a leading 39.4\%.

\paragraph{Scaling and cross-family generalization}
Beyond the main Qwen3-8B result trained on the full $D_{\text{train}}$ corpus, we evaluate the same recipe at a matched 20K Syn+Pub Balanced budget on Qwen3 (0.6B-14B) and on Gemma-3-4B-it and Llama-3.2-3B-Instruct. Gains remain positive throughout, with the strongest headline Strict Precision improvements at 8B (+25.8 / +23.2\,pp on single-hop / single-turn SP); the 8B proxy multi-hop SP (21.0\%) is not comparable to the main 80.7\% in Table~\ref{tab:main_results}. The two non-Qwen backbones gain +33.2 / +31.4\,pp and +33.6 / +32.5\,pp on single-hop / single-turn SP (Appendix~\ref{sec:scaling_laws}).

\subsection{Ablations} 
\label{sec:ablations}
The synthetic pipeline supplies missing structure rather than replacing the scale and domain coverage of public corpora. The additive effect of the 2.9K synthetic subset on merged public data is reported at a 20K budget in Table~\ref{tab:proxy_ablations}. Ablations I and II isolate structural distribution (public versus synthetic profiles, then mixed versus single-complexity curricula), while Ablation III tests whether Anchor Linkage is required for multi-turn coherence. Candidate-pool size and unseen-tool generalization are reported in Appendix~\ref{sec:exp_details}.

\paragraph{Ablation I: Public vs.\ synthetic structure}
Evaluated on the global split (Table~\ref{tab:ablation_composition}), the public subset scores well from coverage, but its serial-to-parallel ratio is skewed (1:5.7 versus 1:1.9 in synthetic), indicating a surplus of independent, one-shot invocations. We isolate this bias on BFCL v3 (Figure~\ref{fig:bfcl_ratio}): the filtered multi-hop split has 141 parallel but only 3 purely serial instances, so strictly sequential trajectories are scarce in public benchmarks. We still stratify on those serial anchors to visualize how scores change as the mix becomes more parallel: serial steps remain harder, and Synthetic Mixed is comparatively stronger on the sequential side, while Public Mixed catches up as tasks flatten. Because the serial subset is small, we read the figure as illustrating public serial scarcity rather than as a powered ranking; the composition claim (synthetic data as a complement for sequential structure, not a substitute for public scale) is identified in Table~\ref{tab:ablation_composition}.

\paragraph{Ablation II: Mixing synthetic complexities}
We next compare specialized single-structure subsets against a mixed synthetic curriculum. All synthetic variants beat vanilla, yet training on one structure mainly helps that category: pure multi-turn has the strongest multi-turn FP (44.4\%) but weaker multi-hop SP, whereas Synthetic Mixed leads multi-hop SP (57.1\%; Table~\ref{tab:ablation_composition}). Figure~\ref{fig:bfcl_ratio} is consistent with this mix-versus-specialization pattern on BFCL v3, with the same serial-subset caveat: Pure Multi-Hop is competitive only when tasks are highly parallel, and Pure Single-Hop lags across the spectrum. Specializing in a single interaction pattern therefore limits transfer; mixing extraction-style single-hop tasks with sequential routing yields more even generalization than any homogeneous subset.

\begin{figure}[t]
    \centering
    \includegraphics[width=0.9\columnwidth]{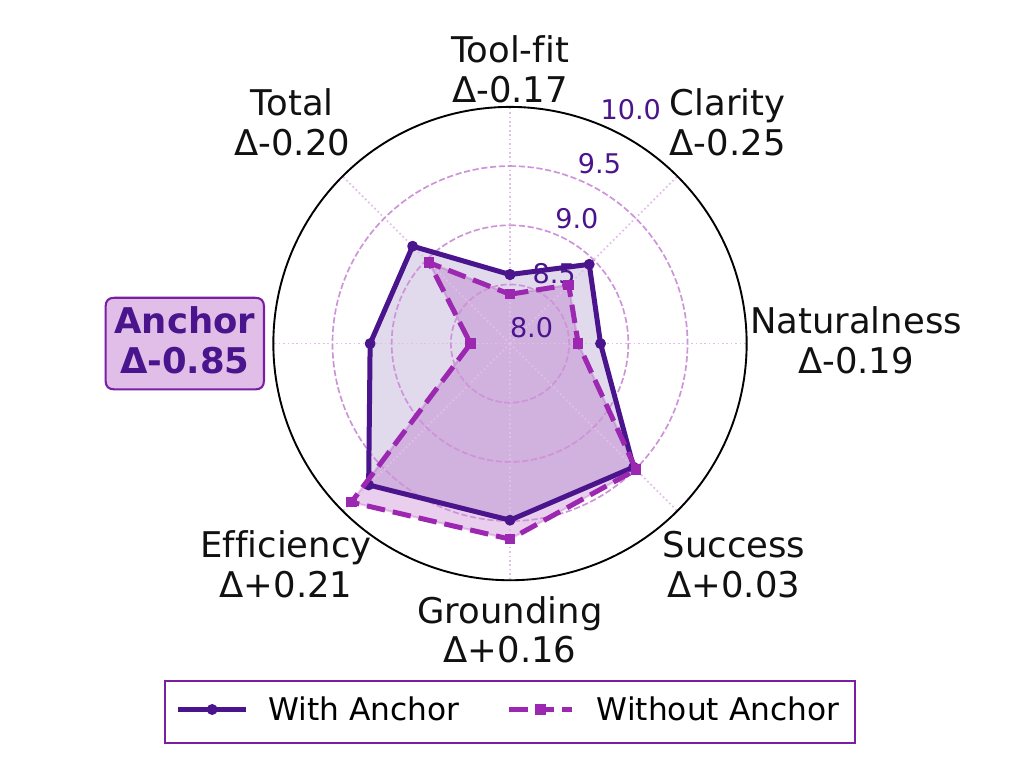} 
    \caption{Intrinsic data quality evaluation for the Anchor Linkage mechanism. Deltas ($\Delta$) indicate the score reduction or increase when the mechanism is removed.}
    \label{fig:radar_anchor}
\end{figure}

\paragraph{Ablation III: Efficacy of the Anchor Linkage Mechanism} 
We sample 10 multi-turn trajectories with and without the Anchor constraint, scored by our LLM rubric (Appendix~\ref{sec:synthetic_eval}). Figure~\ref{fig:radar_anchor} shows only small changes ($-0.25$ to $+0.21$) on generic metrics, because unconstrained generators still produce fluent, locally coherent turns that those scores do not penalize. The \textit{Anchor} dimension, however, drops sharply ($\Delta -0.85$): later queries fail to inherit identifiers and returned values from turn $t-1$, replacing them with locally plausible but ungrounded slots. Appendix~\ref{sec:appendix_anchor_failures} shows the corresponding failure modes, including severe topic drift (Anchor 2/10) and weaker consecutive-turn breakage despite occasional entity reuse (Anchor 5/10). These cases illustrate why a dedicated Anchor score is required: Clarity, Naturalness, and Success remain high even when turn $t$ no longer depends on turn $t-1$. Anchor Linkage is therefore a necessary constraint for synthesizing coherent multi-turn data, rather than a check that generic fluency metrics would already provide.

\section{Conclusion}
\label{sec:conclusion}

In this paper, we presented \method, a unified framework for tool learning in LLM agents. Our framework standardizes the entire pipeline from toolset construction and hybrid data synthesis to evaluation under a shared QAOA representation. By integrating large-scale public corpora with structurally controlled synthetic trajectories, the resulting training dataset contains 390K+ instances covering diverse interaction patterns, including single-hop, multi-hop, single-turn and multi-turn scenarios with both serial and parallel execution structures. In addition, we construct a unified benchmark that enables fine-grained evaluation across function-call, turn, and conversation levels. Experiments show that models trained with our framework achieve strong improvements in tool selection and parameter generation, highlighting the importance of explicitly modeling structural diversity for robust function-calling prediction. While this work provides a standardized foundation for offline tool-use evaluation, we plan to extend the framework to dynamic agent interactions and evaluate long-horizon behaviors in real-world environments with live tool execution in future work.
\section*{Limitations}
\label{sec:limitation}

Due to computational constraints, our experiments use a maximum context length of 8192 tokens, which restricts extremely long-horizon interactions and large tool outputs. Our evaluation prioritizes large-scale standardization over live interactive execution. To enable fair, multi-granularity comparison, we convert benchmarks into unified QAOA and score with static ground-truth matching; this does not capture real-time error recovery or policy correction, and conversion may drop idiosyncratic features of the native sources. Relatedly, the current protocol scores fixed-prompt tool-call generation. Interactive multi-turn parameter filling, where missing arguments are elicited across follow-ups (as in HammerBench multi-turn), is a natural future extension of QAOA rather than a capability mixed into $D_{\text{test}}$.

\section*{Ethics Statement}
\label{sec:ethics} 

For the integration of public data, we exclusively used open-source datasets that have been previously released under permissive licenses. During the synthetic data generation process, our prompting mechanisms and LLM-based planners were explicitly instructed to simulate fictitious user intents and generic business scenarios. We confirm that no personally identifiable information or sensitive user data was scraped, generated, or included in our final dataset. Furthermore, while our data synthesis relies on LLMs, which may inherently reflect societal biases, our multi-stage quality filtering and strict argument-grounding rubrics significantly mitigate the risk of generating unsafe or hallucinated content. All scientific artifacts, including base models and MCP server definitions, were used strictly in accordance with their intended purposes and licenses. Therefore, we believe that our research complies with the ACL Code of Ethics. We used ChatGPT and Gemini for minor language polishing and grammar correction. All technical content, experiments, and conclusions were generated and verified by the authors.


\bibliography{custom}

\clearpage
\appendix

\section*{Appendix}
\section{Dataset details} 
\label{sec:appendix_dataset_details}

\begin{table*}[htbp]
\centering
\small
\resizebox{\textwidth}{!}{%
\begin{tabular}{llll}
\toprule
\multicolumn{4}{c}{\textbf{Servers from mcp.so (Top 40)}} \\
\midrule
302\_browser\_use\_mcp & 302\_sandbox\_mcp & agentql-mcp-server & amap-maps \\
aws-kb-retrieval-server & baidu-map & blender & brave-search \\
context7 & devcontext & edgeone-pages-mcp & everart \\
fetch & firecrawl-mcp-server & framelink-figma-mcp-server & github \\
gitlab & google-maps & howtocook-mcp & jina-ai-mcp-tools \\
mailtrap-email-sending-mcp & mcp-advisor & mcp-server-flomo-mcp-server & minimax-mcp \\
neon-mcp-server & notion-mcp-server & perplexity-ask-mcp-server & playwright-mcp \\
postgresql & puppeteer & qiniu-mcp-server & redis \\
search1api & sentry & sequential-thinking & serper-mcp-server \\
slack & time & todoist-mcp & zhipu-web-search \\
\midrule
\multicolumn{4}{c}{\textbf{Servers from MCP-Universe (11 Servers)}} \\
\midrule
blender & calculator & date & fetch \\
github & google-maps & google-search & notion \\
playwright & weather & yfinance & \\
\bottomrule
\end{tabular}%
}
\caption{The complete list of collected MCP servers used in our toolset construction.}
\label{tab:mcp_servers_list}
\end{table*}

\subsection{Toolset construction}
\label{sec:toolset}

As illustrated in Figure~\ref{fig:toolset}, the toolset is formally defined as the union of six distinct subsets drawn from three primary sources: (1) Academic benchmarks: We integrated tools from established benchmarks to ensure comparability. This includes \textsc{FC-RewardBench} ($T_{\text{fc}}$)~\cite{agarwal2025toolrm} and \textsc{ToolRet-train} ($T_{\text{ret}}$)~\cite{shi2025retrieval}. (2) MCP servers: To capture real-world tool usage patterns, we collected Model Context Protocol (MCP) servers. This subset comprises the top 40 servers listed on mcp.so at the time of collection ($T_{\text{so}}$)~\cite{mcpso_web} and 11 servers utilized in \textsc{MCP-Universe} ($T_{\text{uni}}$)~\cite{servers2025mcp}. The specific list of MCP servers is provided in Table~\ref{tab:mcp_servers_list}. (3) Constructed datasets: This subset includes the specific tool definitions extracted from the training ($T_{\text{train}}$) and test ($T_{\text{test}}$) datasets constructed in this study. 

\subsection{Tool classification taxonomy}
\label{sec:appendix_taxonomy}

\paragraph{Functional Categories}
Based on common API usage patterns observed in agent systems, we define six functional categories:

\begin{itemize}
\item \textbf{Analysis:} Data analysis and insights (statistical analysis, trend analysis, data mining, predictive analysis, business intelligence, etc.)
\item \textbf{Operations:} Business process operations (create, update, delete, workflow management, business logic execution, etc.)
\item \textbf{System:} System administration and maintenance (system configuration, user management, system monitoring, technical maintenance, etc.)
\item \textbf{Visualization:} Data visualization and presentation (chart generation, report creation, data display, dashboard creation, etc.)
\item \textbf{Search:} Information retrieval and search (full-text search, fuzzy search, index query, structured query, data lookup, etc.)
\item \textbf{Generate:} Content and data generation (content generation, code generation, intelligent recommendation, AI generation, automated creation, etc.)
\end{itemize}

\paragraph{Application domains}
Tools are further associated with one of thirteen application domains to reflect real-world usage scenarios:

\begin{itemize}
\item \textbf{Finance:} Finance related (payment, investment, wealth management, insurance, trading, etc.)
\item \textbf{Technology:} Technology and software development (programming, system management, software tools, IT infrastructure, etc.)
\item \textbf{Education:} Education and learning (academic courses, training programs, educational content, learning management, etc.)
\item \textbf{Healthcare:} Medical and health services (medical treatment, health monitoring, medical devices, healthcare management, etc.)
\item \textbf{Entertainment:} Entertainment and media (music, games, film/TV, social entertainment, news, content creation, etc.)
\item \textbf{Travel:} Travel and transportation (tourism, transportation, accommodation, attractions, travel planning, etc.)
\item \textbf{Business:} Business management (enterprise operations, marketing, customer relations, business processes, etc.)
\item \textbf{Lifestyle:} Daily life services (shopping, food, housekeeping, personal tools, consumer services, etc.)
\item \textbf{Science:} Scientific research and analysis (research projects, scientific experiments, academic studies, data analysis, etc.)
\item \textbf{Social:} Social communication and community (social networking, communication tools, community management, collaboration, etc.)
\item \textbf{Sports:} Sports and fitness (sports activities, fitness training, sports events, athletic performance, etc.)
\item \textbf{Environment:} Environment and sustainability (environmental protection, climate monitoring, ecology, sustainable development, etc.)
\item \textbf{Culture:} Culture and arts (art, literature, history, cultural events, language learning, creative content, etc.)
\end{itemize}

\subsection{Toolset filtering details}
\label{sec:appendix_tool_filtering}

\begin{figure*}[htbp]
\centering 
\includegraphics[width=\textwidth]{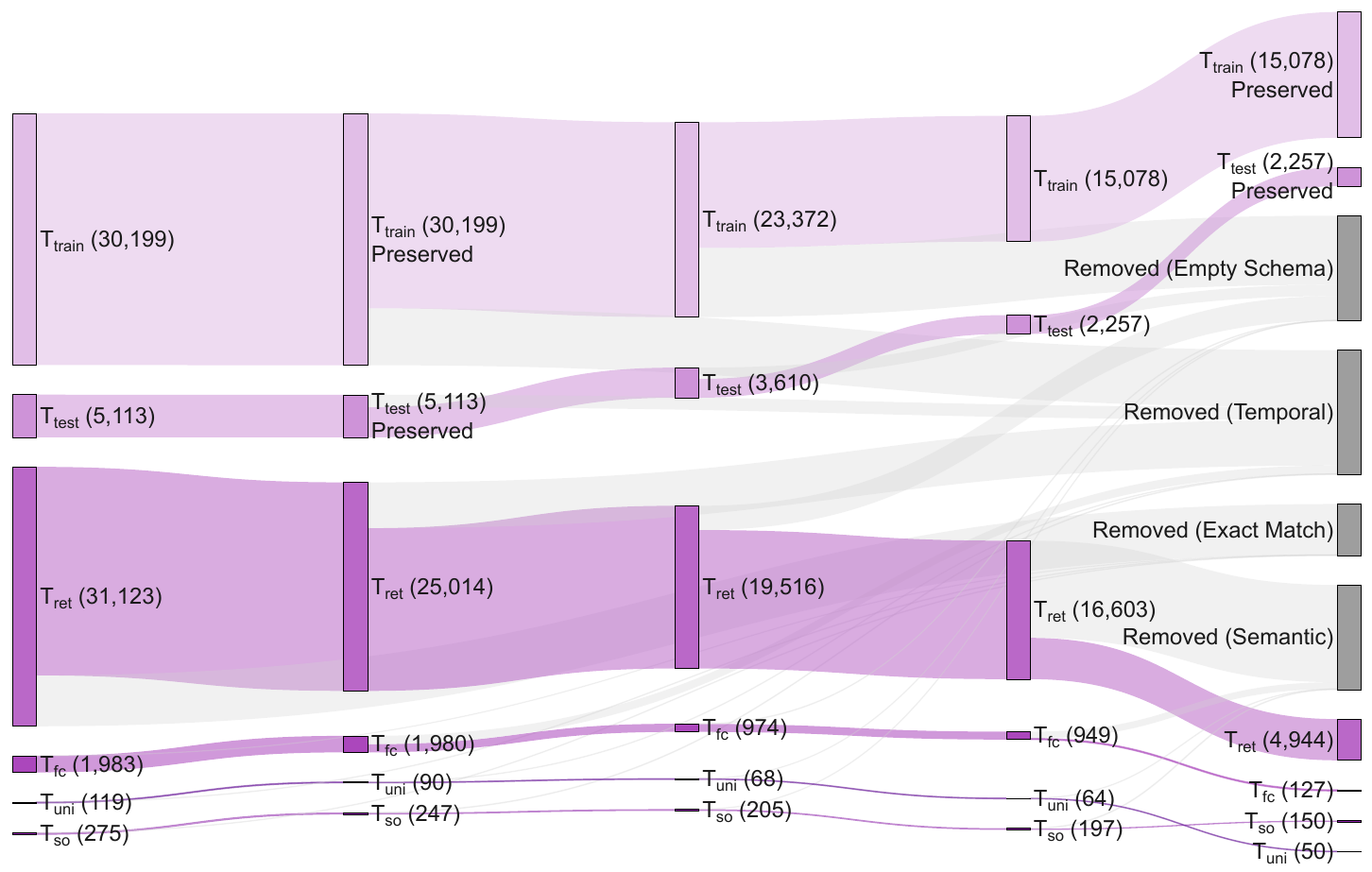} 
\caption{The multi-stage data reduction flow of our toolset quality filtering process. Gray indicates the tool being deduplicated, and purple indicates the tool being retained.} 
\label{fig:toolset} %
\end{figure*}

Because the collected tools originate from heterogeneous sources, the raw pool contains redundancy and incomplete definitions. As illustrated in Figure~\ref{fig:toolset}, we therefore apply a multi-stage filtering pipeline to improve tool quality and ensure fair evaluation. First, we remove exact duplicates within and across subsets based on tool names and descriptions. Second, we exclude tools whose schemas rely on temporal attributes, since different benchmarks adopt inconsistent conventions for resolving relative time expressions (Appendix~\ref{sec:appendix_time_keywords}). Third, we discard tools with missing or invalid parameter schemas to ensure that each tool provides sufficient information for argument generation. Finally, we perform semantic deduplication using embedding similarity to remove functionally redundant tools with different names.

\paragraph{Exact deduplication}
Tools extracted from public datasets often contain duplicates because identical tool definitions appear in multiple query-tool pairs. 
We first perform intra-subset deduplication by removing entries with identical tool names and descriptions. 
This is followed by inter-subset deduplication across different tool sources. 
To preserve dataset consistency, tools belonging to $T_{\text{train}}$ and $T_{\text{test}}$ are retained even when duplicates are detected across external subsets.

\paragraph{Schema validation}
We remove tools with incomplete definitions, such as those lacking a valid parameter schema. 
Tools that contain only a name or description without argument specifications cannot provide sufficient supervision for learning the mapping between user queries and structured tool arguments.

\paragraph{Semantic deduplication}
To identify semantically redundant tools with different names, we encode the concatenation of each tool's name and description using \textsc{Qwen3-Embedding-8B}. 
Cosine similarity is computed using FAISS. 
Tools with similarity greater than 0.9 are considered duplicates. 
When duplicates are detected, instances from external subsets are removed while those belonging to $T_{\text{train}}$ and $T_{\text{test}}$ are retained to preserve dataset consistency.

\subsection{Public data quality evaluation and filtering}
\label{sec:public_summary}

To construct a structurally consistent training corpus, we apply uniform filtering principles across public training datasets. Table~\ref{tab:public_summary} summarizes the dataset-level normalization and filtering applied to $D_{\text{pub}}$ prior to integration. Across these corpora, we enforce the following dataset-agnostic criteria:

\begin{itemize}
    \item \textbf{Schema completeness}: each sample must contain a well-formed user query, a valid tool call (or candidate API schema), and, when applicable, observations and final answers.
    \item \textbf{Executable supervision}: we discard items with missing function calls, incomplete parameter specifications, invalid JSON structure, or empty/malformed ground-truth traces.
    \item \textbf{Language normalization}: only English-language user queries and assistant messages are retained.
    \item \textbf{Toolset compatibility}: samples referencing tools removed during tool filtering (e.g., temporal-sensitive or redundant tools) are excluded.
    \item \textbf{Invalid-category filtering}: subsets explicitly marked as irrelevant or lacking actionable ground truth are removed.
\end{itemize}

Field-level conversion decisions and sample retention for the seven evaluation benchmarks in $D_{\text{test}}$ are detailed separately in Appendix~\ref{sec:benchmark_conversion}.

\begin{table*}[t]
\centering
\small
\begin{tabularx}{\textwidth}{@{}l X@{}}
\toprule
\textbf{Dataset} & \textbf{Processing Summary} \\
\midrule

API-Bank~\cite{li2023apibank} &
Filtered for training use; standardized tool schemas and function-call traces into QAOA. \\[4pt]

ToolAlpaca~\cite{tang2023toolalpaca} &
Transformed API descriptions and queries into single-hop QAOA format; removed structurally inconsistent items. \\[4pt]

ToolHop~\cite{ye2025toolhop} &
Retained items convertible to multi-hop trajectories; removed unresolved-hop subsets; normalized argument formats and tool identifiers. \\[4pt]

APIGen~\cite{liu2024apigen} &
Filtered structurally invalid entries from 60{,}000 raw items; standardized tool schemas; integrated valid samples into QAOA. \\[4pt]

Seal-Tools~\cite{wu2024sealtools} &
Preserved train/dev partitions for training; standardized tool schemas; converted entries into QAOA with explicit tool definitions. \\[4pt]

Toucan~\cite{xu2025toucan} &
Converted 1.37M raw samples; removed incomplete or structurally inconsistent trajectories; retained QAOA-normalized conversations. \\[4pt]

Tool-calling~\cite{hermes_tool_use} &
Unified formatting into QAOA; removed schema-mismatched or incomplete entries; retained samples with valid function-call traces. \\[4pt]

Junaidjk~\cite{junaidjk_fc} &
Unified formatting into QAOA; removed schema-mismatched or incomplete entries; retained samples with valid function-call traces. \\[4pt]

Vikhrmodels~\cite{vikhrmodels_tool} &
Standardized tool schemas; resolved formatting inconsistencies; retained entries convertible to well-formed function calls. \\[4pt]

MathAndMagic~\cite{mathandmagic_fc} &
Normalized function-calling traces; removed incomplete or invalid entries; retained consistent QAOA-formatted samples. \\[4pt]

\bottomrule
\end{tabularx}
\caption{Public training-data quality evaluation and filtering.}
\label{tab:public_summary}
\end{table*}

\begin{table}[t]
\centering
\small
\resizebox{\columnwidth}{!}{%
\begin{tabular}{lrr}
\toprule
\textbf{Dataset} & \textbf{Conv.} & \textbf{Filt.} \\
\midrule
\multicolumn{3}{c}{\textit{Training Data ($D_{\text{pub}}$)}} \\
\midrule
API-Bank~\cite{li2023apibank} & 338 & 122 \\
ToolAlpaca~\cite{tang2023toolalpaca} & 4,096 & 2,429 \\
ToolHop~\cite{ye2025toolhop} & 995 & 7 \\
APIGen~\cite{liu2024apigen} & 60,000 & 28,666 \\
Seal-Tools~\cite{wu2024sealtools} & 12,022 & 5,214 \\
Toucan~\cite{xu2025toucan} & 1,367,983 & 319,669 \\
Tool-calling~\cite{hermes_tool_use} & 35,786 & 8,692 \\
Junaidjk~\cite{junaidjk_fc} & 13,850 & 3,470 \\
Vikhrmodels~\cite{vikhrmodels_tool} & 3,396 & 2,493 \\
MathAndMagic~\cite{mathandmagic_fc} & 22,218 & 16,361 \\
\cmidrule(lr){1-3}
\textbf{Total (Train)} & \textbf{1,520,684} & \textbf{387,123} \\
\midrule
\multicolumn{3}{c}{\textit{Evaluation Benchmark ($D_{\text{test}}$)}} \\
\midrule
BFCL v3~\cite{patil2025bfcl} & 3,065 & 984 \\
ACEBench~\cite{chen2025acebench} & 250 & 59 \\
Seal-Tools~\cite{wu2024sealtools} & 1,354 & 579 \\
HammerBench~\cite{wang2025hammerbench} & 6,531 & 4,340 \\
ComplexFuncBench~\cite{zhong2025complexfuncbench} & 1,000 & 51 \\
API-Bank~\cite{li2023apibank} & 50 & 35 \\
ToolAlpaca~\cite{tang2023toolalpaca} & 209 & 115 \\
\cmidrule(lr){1-3}
\textbf{Total (Test)} & \textbf{12,459} & \textbf{6,163} \\
\bottomrule
\end{tabular}%
}
\caption{Statistics of the public datasets integrated into our framework. Conv. (Converted Count) represents the initial number of conversations obtained after standardizing the raw heterogeneous data into our unified QAOA format. Filt. (Filtered Count) indicates the final retained size after our rigorous quality filtering mechanism.}
\label{tab:public_datasets}
\end{table}
\begin{figure*}[t]
    \centering
    \includegraphics[width=\textwidth]{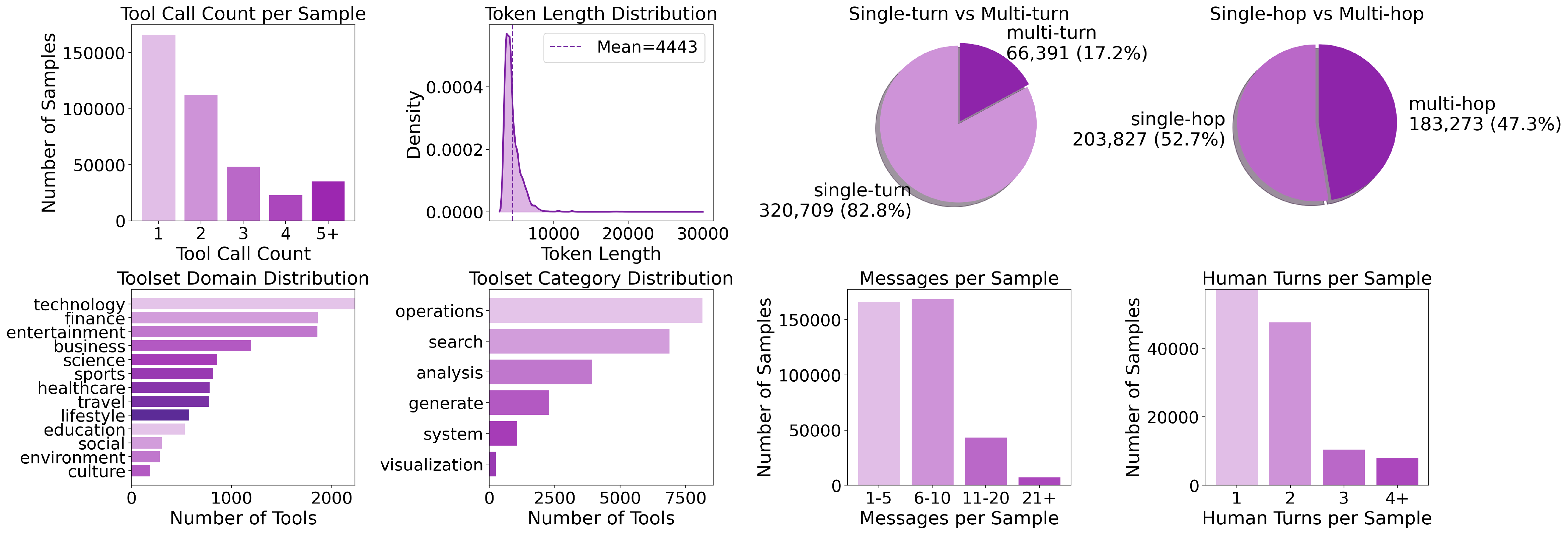} 
    \caption{Comprehensive statistics of our unified training dataset $D_{\text{Pub}}$. The top row illustrates the structural complexity and scale, including the distribution of tool calls per sample, token length density, and the proportions of multi-turn and multi-hop trajectories. The bottom row demonstrates the broad semantic diversity across tool domains and functional categories, alongside conversation density metrics. Note that Messages per Sample reflects the total count of user queries, tool calls, environment observations, and assistant answers within a single dialogue episode.}
    \label{fig:data_stats}
\end{figure*}

After standardization, we further filtered samples based on the finalized training and evaluation tool inventories, removing conversations that referenced tools excluded during toolset filtering. The statistics are illustrated in Table~\ref{tab:public_datasets}.

\subsection{Evaluation benchmark conversion checklist}
\label{sec:benchmark_conversion}

Converting seven heterogeneous public benchmarks into a shared QAOA schema involves two complementary steps that are easy to conflate: (1)~\textbf{Field-level} decisions on what information is preserved, normalized, or omitted from the QAOA slots; and (2)~\textbf{Sample-level} rules for which official evaluation instances enter $D_{\text{test}}$. Our protocol targets fixed-prompt tool-call generation: given the user utterance(s) and candidate tools, does the model emit the correct gold tool name(s) and arguments? We therefore include only samples with a concrete, checkable gold function call under this setting. Official subsets that evaluate a different capability (e.g., irrelevance detection without a unique gold call, or interactive slot-filling where the user must supply missing arguments across turns, as in HammerBench multi-turn and API-Bank Level-1/2) are documented in Table~\ref{tab:benchmark_sample_retention} but lie outside the scope of $D_{\text{test}}$.

Table~\ref{tab:benchmark_field_conversion} summarizes field-level conversion decisions. Table~\ref{tab:benchmark_sample_retention} reports post-conversion (Conv.) versus raw counts per benchmark, together with the main out-of-scope subsets; a subsequent tool-inventory filter yields the Filt.\ sizes in Table~\ref{tab:public_datasets}, which are the counts used in $D_{\text{test}}$. Normalization unifies surface formats (roles, system prompts, schema keys) without changing what counts as a correct tool call: gold tool names and arguments remain the basic scoring targets under our shared matcher (Appendix~\ref{sec:appendix_matching_rules}).

\begin{table*}[t]
\centering
\small
\begin{tabularx}{\textwidth}{@{}l X@{}}
\toprule
\textbf{Information} & \textbf{Conversion decision} \\
\midrule
User query / instruction &
Preserved: kept as the human turn. \\[3pt]

Gold tool name(s) and arguments &
Preserved: kept as the scored function call. \\[3pt]

Tool name, description, argument schema &
Preserved: candidate schemas retained; parameters renamed to inputSchema. \\[3pt]

Dialogue roles (user, assistant, \dots) &
Normalized: remapped to QAOA roles (human, function call, observation, gpt). \\[3pt]

System / instruction prompt &
Normalized: replaced by the unified UniToolCall system prompt. \\[3pt]

Final natural-language answer &
Preserved: gpt slot kept; content retained when present in raw, otherwise empty, not a scored target. \\[3pt]

Tool return / observation &
Preserved: real returns kept for ComplexFuncBench and API-Bank; often empty elsewhere. \\[3pt]

Samples with no unique gold tool call &
Not included: e.g., irrelevance-only splits, preference-only answers, missing-tool turns. \\[3pt]

Interactive slot-filling multi-turn &
Not included: user must supply missing arguments via follow-ups (Hammer MT; API-Bank L1/2). \\[3pt]

Non-English queries &
Not included: English-only evaluation. \\
\bottomrule
\end{tabularx}
\caption{Field-level benchmark conversion decisions for the seven evaluation benchmarks. Conversion unifies surface formats while retaining the scorable core (query, gold call, tool schema).}
\label{tab:benchmark_field_conversion}
\end{table*}

\begin{table*}[t]
\centering
\small
\begin{tabularx}{\textwidth}{@{}l X@{}}
\toprule
\textbf{Benchmark} & \textbf{Retention summary} \\
\midrule
BFCL v3~\cite{patil2025bfcl} &
Retained 3{,}065 of 4{,}441 raw samples. Out of scope: irrelevance and no-gold-call splits, multi\_turn\_miss\_func, non-English queries, and empty ground truth. \\[3pt]

ACEBench~\cite{chen2025acebench} &
Retained 250 single-turn tool-call samples (250 raw in the selected core). Out of scope: atom-type, agent, preference, and interactive multi-turn subsets. \\[3pt]

Seal-Tools~\cite{wu2024sealtools} &
Retained all 1{,}354 official test samples. Train and validation splits are outside $D_{\text{test}}$. \\[3pt]

HammerBench~\cite{wang2025hammerbench} &
Retained 6{,}531 of 13{,}054 English single-turn samples. Out of scope: irrelevance candidates, Chinese subsets, and multi-turn slot-filling dialogues. \\[3pt]

ComplexFuncBench~\cite{zhong2025complexfuncbench} &
Retained all 1{,}000 samples; format normalized only. \\[3pt]

API-Bank~\cite{li2023apibank} &
Retained all 50 Level-3 samples. Out of scope: Level-1/2 slot-filling dialogues; ToolSearcher plan/retrieve steps stripped from Level-3. \\[3pt]

ToolAlpaca~\cite{tang2023toolalpaca} &
Retained 209 of 214 instructions; five items dropped because the gold action is absent from the parsed tool list. \\
\bottomrule
\end{tabularx}
\caption{Sample retention after converting the seven public evaluation benchmarks into QAOA.}
\label{tab:benchmark_sample_retention}
\end{table*}

\subsection{Synthetic data quality evaluation and filtering}
\label{sec:synthetic_eval}

To ensure the quality and consistency of the synthetically generated dataset without relying on external proprietary models, we design a unified LLM-based self-evaluation framework. The generator model (\textsc{Qwen3-32B}) evaluates its own generated QAOA trajectories across a set of fine-grained metrics. The framework is shared across single/multi-hop and single/multi-turn datasets, with minor extensions for multi-turn episodes.

\subsubsection*{Evaluation dimensions}

The evaluation rubric consists of six core metrics grouped into two dimensions, plus an additional episode-level metric for multi-turn data. Each metric is scored on a scale from 1 to 10 by the generator model.

\noindent\textbf{Query evaluation:} This dimension evaluates the initial user query $q$ with respect to the available tools:

\begin{itemize}
    \item \textbf{Tool-fit:} Whether the query is appropriately designed around the available tool capabilities and implicitly or explicitly provides the necessary parameters.
    \item \textbf{Clarity:} Whether the task specification is unambiguous, well-defined, and provides sufficient constraints for planning a valid solution.
    \item \textbf{Naturalness:} Whether the query resembles a realistic user request in a practical scenario rather than a templated or system-style prompt.
\end{itemize}

\noindent\textbf{Trajectory evaluation:} This dimension evaluates the correctness and coherence of the generated trajectory consisting of Action ($a$), Observation ($o$), and Answer ($r$):

\begin{itemize}
    \item \textbf{Success:} Whether the generated tool calls and final answer successfully complete the user's task.
    \item \textbf{Grounding:} Whether the final response is strictly supported by the simulated observations, without hallucinated facts or inconsistent parameters.
    \item \textbf{Efficiency:} Whether the trajectory completes the task using a concise and non-redundant sequence of tool calls.
\end{itemize}

\noindent\textbf{Anchor linkage:} For multi-turn episodes, we extend the rubric with one additional episode-level metric:

\begin{itemize}
    \item \textbf{Anchor Linkage ($s_{\text{anchor}}$):} Measures whether later turns explicitly and consistently reference anchors introduced in previous turns, and whether such references are functionally meaningful for subsequent tool usage.
\end{itemize}

\subsubsection*{Acceptance thresholds}

For all synthetically generated candidates that pass basic schema and formatting checks, we apply strict acceptance criteria based on the evaluation scores.

\noindent\textbf{Single-hop and multi-hop instances:} A trajectory is accepted only if the following conditions are simultaneously satisfied:

\begin{itemize}
    \item \textbf{Minimum score constraint:} The lowest score among all six metrics must be at least $4.0$ ($\min(S) \ge 4.0$).
    \item \textbf{Average score constraint:} The average score across the six metrics must be at least $8.0$ ($\text{avg}(S) \ge 8.0$).
\end{itemize}

These constraints ensure that no individual dimension is critically flawed while maintaining high overall quality.

\noindent\textbf{Multi-turn episodes:} For multi-turn data, acceptance is determined by a weighted comprehensive score:

\begin{equation}
\begin{aligned}
S &= 0.4\,\text{Query}_{\text{avg}}
   + 0.4\,\text{Trajectory}_{\text{avg}} \\
  &\quad + 0.2\,s_{\text{anchor}} \\
\end{aligned}
\end{equation}

A multi-turn episode is accepted only if $S \ge 8.0$ and the minimum score across all dimensions is at least $4.0$.

\noindent\textbf{Self-Refinement loop:} If a generated trajectory fails to satisfy the above criteria, the pipeline triggers an automatic self-correction loop.
The generator model is instructed to regenerate the trajectory for the same target tool, with a maximum of three retries.
If no valid trajectory is produced after all attempts, the corresponding tool is excluded from the synthetic dataset $D_{\text{syn}}$.

\subsection{Statistics of \texorpdfstring{$D_{\text{Pub}}$}{D\_Pub}}
\label{sec:other_details}
Figure~\ref{fig:data_stats} shows the comprehensive statistics of our unified training dataset $D_{\text{Pub}}$. Collectively, these distributions highlight the dataset's high diversity and rigorous complexity.

\subsection{Implementation Details of Anchor Linkage}
\label{sec:appendix_anchor_linkage}

To robustly implement the Anchor Linkage mechanism without suffering from high trajectory generation failure rates, we design a three-stage hybrid pipeline within our data synthesis engine:

\begin{itemize}
    \item \textbf{Dynamic Prompting Constraint:} When generating the user query for turn $t$, the generator LLM is initially prompted to organically incorporate required\_anchors (e.g., transaction IDs or specific parameters returned in the observation at turn $t-1$). If the initial generation attempt fails to include the anchor, a hard constraint (e.g., Hard constraint: the user message must explicitly include...) is dynamically injected into the prompt for subsequent retries.
    
    \item \textbf{Programmatic Verification and Fallback:} We apply a regex-based programmatic verification step to check for the exact string presence (or boolean/numeric equivalence) of the anchor within the generated query. If the LLM consistently fails after the maximum number of retries, an automated programmatic fallback is triggered. It forcefully prepends the anchor to the user query (e.g., About [anchor], ...) to guarantee state inheritance and prevent the pipeline from collapsing.
    
    \item \textbf{Evaluation Gating:} Finally, textual inclusion alone does not guarantee logical coherence. The entire trajectory must pass the $s_{\text{anchor}}$ LLM-evaluation rubric (detailed in Appendix~\ref{sec:synthetic_eval}) as a final gate. This ensures that the inherited anchor is functionally meaningful for the subsequent tool invocations. Trajectories failing this quality check are completely discarded.
\end{itemize}

\subsection{Qualitative failures without Anchor Linkage}
\label{sec:appendix_anchor_failures}

Our with-/without-anchor ablation regenerates user queries under each condition rather than holding queries fixed; Cases A and B below illustrate failure modes under unconstrained generation. Case C shows a separate episode generated with the constraint (not paired with A/B). Aggregate Anchor scores drop substantially when the constraint is removed (Figure~\ref{fig:radar_anchor}).

\subsubsection*{Case A (without constraint; Anchor 2/10; lifestyle, 3 turns)}

\begin{itemize}
    \item \textbf{Scenario plan:} organize lifestyle services after moving into a new apartment.
    \item \textbf{Turn 1:} I want to book a big cleaning service for my new apartment. Can you confirm the details for service ID 13 covering 3 rooms?
    \item \textbf{Turn 2:} I need a shopping list for vegetarian meals for two people with a 500 yuan weekly budget. Please format it as a categorized list and include substitute options if any ingredients are unavailable. I already have quinoa, almond milk, and brown rice in my pantry, so avoid duplicating those.
    \item \textbf{Turn 3:} I want to update my app settings. My user ID is user\_12345. Turn off marketing notifications and enable dark mode for the theme preference. Also, make sure SMS notifications are disabled but email alerts stay on.
    \item \textbf{Linkage summary:} Turn 1$\rightarrow$2: broken (cleaning $\rightarrow$ shopping; service ID 13 and 3 rooms not carried over). Turn 2$\rightarrow$3: broken (shopping $\rightarrow$ app settings; budget and pantry items not carried over).
\end{itemize}

\subsubsection*{Case B (without constraint; Anchor 5/10; technology, 4 turns)}

\begin{itemize}
    \item \textbf{Scenario plan:} secure location sharing with authentication integration.
    \item \textbf{Turn 1:} I need to share my current location at 37.7749, -122.4194 with a pre-registered contact named John Doe. Can you confirm the coordinates are properly formatted and verify this contact exists in the system?
    \item \textbf{Turn 2:} I need to configure the authentication layer using the Google OIDC provider. My client credentials are: client ID 1234567890-abcdefg.apps.googleusercontent.com and secret GOOGLE\_SECRET\_1234. The redirect URI for this integration should be https://app.example.com/auth/callback. Can the system validate these credentials and establish the provider mapping?
    \item \textbf{Turn 3:} I need to test the connectivity to the location-sharing API endpoint at https://api.example.com/v1/share using a POST request. The payload should include my location 37.7749, -122.4194 and contact name John Doe. Please validate the authentication headers by including the Authorization Bearer token from the Google OIDC provider and confirm the response structure with a 200 status code. Set request\_heartbeat to true to verify the system's responsiveness after this test.
    \item \textbf{Turn 4:} I need to extract demographic information from the collected user data records. The data includes entries like \{"name": "John Doe", "age": 30\} and \{"name": "Jane Smith", "age": 25\}. Please apply the personal\_info schema to structure this data for analytics.
    \item \textbf{Linkage summary:} Turn 1$\rightarrow$2: broken (location and John Doe omitted). Turn 2$\rightarrow$3: partial (Turn-1 location and John Doe return, but Turn-2 credential anchors dropped). Turn 3$\rightarrow$4: broken (API endpoint, OIDC token, and location payload not carried over).
\end{itemize}

\subsubsection*{Case C (with constraint; Anchor 10/10; technology, 3 turns)}

\begin{itemize}
    \item \textbf{Scenario plan:} development workspace setup with OIDC authentication and API communication. This episode is not paired with Cases A/B; it illustrates successful linkage under the constraint.
    \item \textbf{Turn 1:} I'm starting to set up my development environment for the SmartAuth project. I need to create a new workspace named SmartAuth-dev using the main branch from the aweme git repository. Can you help me with the initial configuration?
    \item \textbf{Turn 2:} I need to configure the OIDC provider for the SmartAuth-dev workspace. The provider should be GoogleAuth with client ID SmartAuth-app and secret dev-sec-2023. I'll also add redirect URIs like ["http://localhost:3000/callback", "https://dev.smartauth.example/auth"] and specify grant types as ["authorization\_code", "refresh\_token"]. Can you help me map these settings correctly?
    \item \textbf{Turn 3:} I need to test the authentication flow for the SmartAuth-dev workspace using the GoogleAuth OIDC provider. Let me send a POST request to the /api/v1/authorize endpoint with the user credentials: \{"username": "test\_user", "password": "SecurePass123!"\}. I'll set the Authorization header to Bearer dev-sec-2023 and expect a valid JWT token response. Can you confirm the request structure and execution?
    \item \textbf{Linkage summary:} Turn 1$\rightarrow$2: linked (SmartAuth-dev). Turn 2$\rightarrow$3: linked (SmartAuth-dev, GoogleAuth, dev-sec-2023).
\end{itemize}

\subsection{Comparison with related tool-use resources}
\label{sec:related_resources}

Table~\ref{tab:related_resources} situates \method against ToolACE, ToolMind, ToolFlow, and TRAJECT-Bench~\citep{he2025traject}. The first three synthesize training trajectories and evaluate on native leaderboards; they do not convert heterogeneous public resources into a shared QAOA representation or a common multi-granularity metric suite. TRAJECT-Bench instead targets trajectory-level diagnostics and agentic settings, without releasing a training corpus, and focuses on single-turn multi-tool trajectories. \method addresses format/protocol fragmentation: a hybrid 390K+ corpus, explicit SH/MH $\times$ ST/MT factorization with serial/parallel control, Anchor Linkage, and unified evaluation over seven converted benchmarks.

\begin{table*}[t]
\centering
\small
\setlength{\tabcolsep}{6pt}
\newcommand{\ok}{\ensuremath{\checkmark}}
\newcommand{\no}{\ensuremath{\times}}
\newcommand{\pt}{\ensuremath{\triangle}}
\newcolumntype{Y}{>{\centering\arraybackslash}X}
\begin{tabularx}{\textwidth}{@{}>{\raggedright\arraybackslash}p{0.30\textwidth} *{5}{Y}@{}}
\toprule
\textbf{Dimension} & \textbf{ToolACE} & \textbf{ToolMind} & \textbf{ToolFlow} & \textbf{TRAJECT} & \textbf{UniToolCall} \\
\midrule
Shared cross-resource representation & \no & \no & \no & \no & \ok \\
Unified public-benchmark evaluation & \no & \no & \no & \no & \ok \\
Training dataset release & \ok & \ok & \pt & \no & \ok \\
Synthetic data pipeline & \ok & \ok & \ok & \no & \ok \\
Hybrid public-data integration & \no & \ok & \no & \no & \ok \\
Explicit hop control (SH/MH) & \pt & \pt & \pt & \ok & \ok \\
Explicit turn control (ST/MT) & \ok & \ok & \ok & \no & \ok \\
Serial / parallel control & \ok & \pt & \pt & \ok & \ok \\
Anchor-style cross-turn constraint & \no & \no & \no & \no & \ok \\
FC / turn / conversation metrics & \ok/\no/\no & \ok/\pt/\pt & \ok/\no/\no & \ok/\no/\no & \ok/\ok/\ok \\
Trajectory-level diagnostics & \no & \no & \no & \ok & \no \\
Agentic evaluation (ReAct, etc.) & \no & \no & \no & \ok & \no \\
\bottomrule
\end{tabularx}
\caption{Comparison with related tool-use resources. $\checkmark$ = first-class support; $\triangle$ = partial or secondary; $\times$ = absent.}
\label{tab:related_resources}
\end{table*}

\section{Benchmark details}
\label{sec:ben_details}

\subsection{Temporal filtering criteria} 
\label{sec:appendix_time_keywords}

Different benchmarks handle temporal parameters inconsistently: ACEBench resolves relative expressions (e.g., tomorrow) into absolute timestamps, whereas HammerBench keeps the original relative forms. To avoid evaluation bias, we exclude tools whose schemas mention temporal attributes. We match keywords such as date, time, datetime, timestamp; day, hour, minute, second, year, month, week; schedule, duration, period; and start\_time, end\_date, pickup\_time, under snake, kebab, camel, and word-boundary forms.

\subsection{Rule-based matching details}
\label{sec:appendix_matching_rules}
Rule-based matching is implemented as strict exact matching after deterministic normalization. This stage is designed to treat formatting variance as equivalent while preserving hard correctness constraints.

\paragraph{Tool name normalization}
Tool names are normalized by removing punctuation, digits, and separators, and then converting to lowercase. For example, uber.ride and uber\_ride become identical after normalization.

\paragraph{Parameter value normalization}
To robustly compare parameter values across heterogeneous outputs, we apply the following canonicalization rules:
\begin{itemize}
    \item \textbf{Date canonicalization}: date strings in multiple formats (e.g., April 1, 2023, 2023-04-01, and 2023/04/01) are normalized to YYYY-MM-DD.
    \item \textbf{Array parsing}: stringified arrays (e.g., [1, 2, 3]) are parsed into actual arrays before comparison.
    \item \textbf{String normalization}: strings are lowercased, punctuation and articles (a/an/the) are removed, and whitespace is ignored; e.g., A black cat and blackcat are treated as identical.
    \item \textbf{Type casting}: mixed representations of the same value are unified, including numeric string--number equivalence (e.g., "40.7128" and 40.7128).
\end{itemize}

\paragraph{Matching criteria}
After normalization, rule-based matching requires full equality on both normalized tool names and all normalized argument key-value pairs. Such normalization improves evaluation fairness and reproducibility by removing superficial formatting variance while preserving exact semantic correctness constraints.

\section{Experimental details}
\label{sec:exp_details}
\begin{figure*}[t]
    \centering
    \includegraphics[width=\textwidth]{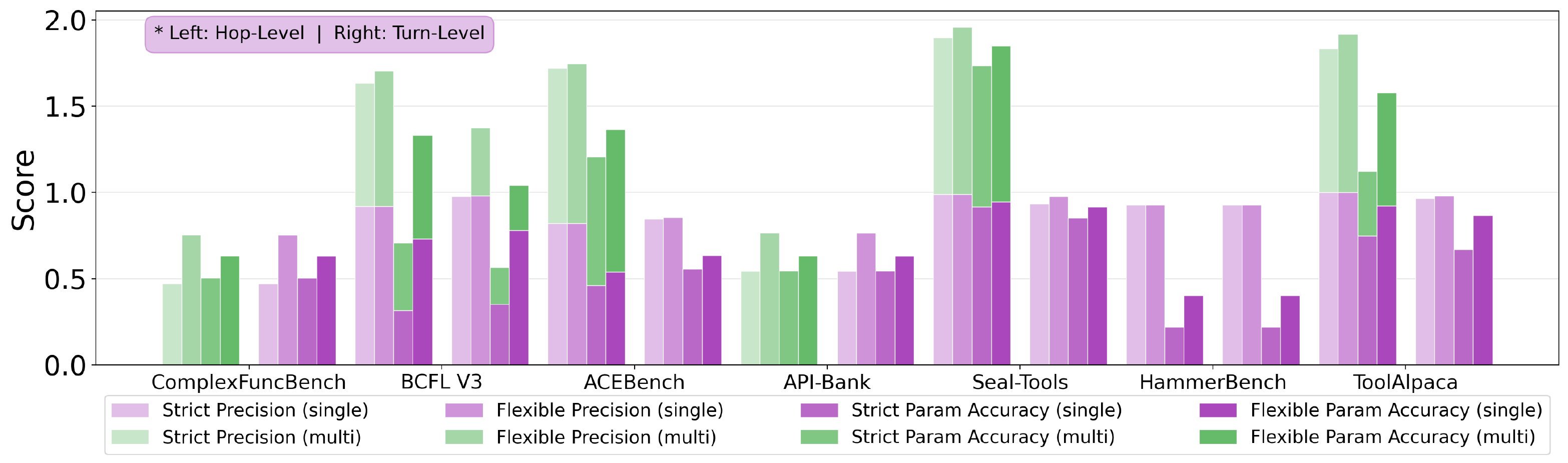} 
    \caption{Performance breakdown of \method across the 7 sub-datasets under the Hybrid-20 setting.}
    \label{fig:subset_results}
\end{figure*}

Figure~\ref{fig:subset_results} breaks down Hybrid-20 performance across the seven evaluation benchmarks.
\begin{table*}[t]
\centering
\small
\resizebox{\textwidth}{!}{%
\begin{tabular}{l cccc cccc}
\toprule
\multicolumn{1}{c}{\multirow{2}{*}{\textbf{Models}}} & \multicolumn{2}{c}{\textbf{SP (\%)~$\uparrow$}} & \multicolumn{2}{c}{\textbf{FP (\%)~$\uparrow$}} & \multicolumn{2}{c}{\textbf{SPA (\%)~$\uparrow$}} & \multicolumn{2}{c}{\textbf{FPA (\%)~$\uparrow$}} \\
\cmidrule(lr){2-3} \cmidrule(lr){4-5} \cmidrule(lr){6-7} \cmidrule(lr){8-9}
 & Single & Multi & Single & Multi & Single & Multi & Single & Multi \\
\midrule
\rowcolor{gray!15} \multicolumn{9}{c}{\textit{\textbf{Hop-Level Scenarios (Single-Hop \& Multi-Hop)}}} \\
\midrule
Qwen3-8B (Vanilla) 
& $69.6 \pm 4.4$ & $24.4 \pm 4.2$ 
& $69.6 \pm 4.4$ & $55.5 \pm 3.9$ 
& $20.8 \pm 1.5$ & $40.1 \pm 4.1$ 
& $38.3 \pm 2.6$ & $49.1 \pm 4.4$ \\
\textbf{\method (Ours)} 
& $93.9 \pm 1.0$ & $80.5 \pm 0.5$ 
& $93.9 \pm 1.0$ & $89.7 \pm 0.2$ 
& $27.4 \pm 0.2$ & $67.0 \pm 0.7$ 
& $49.1 \pm 0.5$ & $78.6 \pm 0.3$ \\
\midrule
\rowcolor{gray!15} \multicolumn{9}{c}{\textit{\textbf{Turn-Level Scenarios (Single-Turn \& Multi-Turn)}}} \\
\midrule
Qwen3-8B (Vanilla) 
& $65.9 \pm 4.4$ & $0.9 \pm 1.6$ 
& $69.2 \pm 4.4$ & $27.0 \pm 2.6$ 
& $23.2 \pm 1.8$ & $12.8 \pm 0.7$ 
& $40.0 \pm 2.8$ & $16.6 \pm 1.2$ \\
\textbf{\method (Ours)} 
& $93.8 \pm 0.8$ & $0.0 \pm 0.0$ 
& $94.6 \pm 0.8$ & $37.8 \pm 1.6$ 
& $31.8 \pm 0.2$ & $16.6 \pm 4.0$ 
& $52.8 \pm 0.4$ & $23.5 \pm 2.3$ \\
\bottomrule
\end{tabular}%
}
\caption{Repeated-run statistics (mean $\pm$ standard deviation) across varying tool-use complexities.}
\label{tab:repeated_run_stability}
\end{table*}
\begin{table*}[t]
\centering
\small
\setlength{\tabcolsep}{3pt} 
\resizebox{\textwidth}{!}{%
\begin{tabular}{l c cccc c cccc c cccc c cccc} 
\toprule
\multicolumn{1}{c}{\multirow{2}{*}{\textbf{Models}}} & & \multicolumn{4}{c}{\textbf{SP (\%)~$\uparrow$}} & & \multicolumn{4}{c}{\textbf{FP (\%)~$\uparrow$}} & & \multicolumn{4}{c}{\textbf{SPA (\%)~$\uparrow$}} & & \multicolumn{4}{c}{\textbf{FPA (\%)~$\uparrow$}} \\
\cmidrule(lr){3-6} \cmidrule(lr){8-11} \cmidrule(lr){13-16} \cmidrule(lr){18-21}
 & & SH & MH & ST & MT & & SH & MH & ST & MT & & SH & MH & ST & MT & & SH & MH & ST & MT \\
\midrule
\rowcolor{gray!15} \multicolumn{21}{c}{\textit{\textbf{Dataset-level Macro-Averaged Results}}} \\
\midrule
Qwen3-8B (Vanilla) & & 71.9 & 31.2 & 45.8 & 0.0 & & 72.0 & 53.6 & 58.7 & 25.8 & & 41.8 & 34.8 & 34.4 & 12.1 & & 56.4 & 45.7 & 47.2 & 16.1 \\
\rowcolor{TrueLightPurple}
\textbf{\method (Ours)} & & \makecell{\textbf{93.2} \\ \textcolor{ForestGreen}{\scriptsize $\uparrow$ 21.3}} & \makecell{\textbf{72.8} \\ \textcolor{ForestGreen}{\scriptsize $\uparrow$ 41.6}} & \makecell{\textbf{81.0} \\ \textcolor{ForestGreen}{\scriptsize $\uparrow$ 35.2}} & \makecell{0.0 \\ \textcolor{WildStrawberry}{\scriptsize $\downarrow$ 0.0}} & & \makecell{\textbf{93.2} \\ \textcolor{ForestGreen}{\scriptsize $\uparrow$ 21.2}} & \makecell{\textbf{85.2} \\ \textcolor{ForestGreen}{\scriptsize $\uparrow$ 31.6}} & \makecell{\textbf{89.2} \\ \textcolor{ForestGreen}{\scriptsize $\uparrow$ 30.5}} & \makecell{\textbf{39.4} \\ \textcolor{ForestGreen}{\scriptsize $\uparrow$ 13.6}} & & \makecell{\textbf{53.2} \\ \textcolor{ForestGreen}{\scriptsize $\uparrow$ 11.4}} & \makecell{\textbf{56.4} \\ \textcolor{ForestGreen}{\scriptsize $\uparrow$ 21.6}} & \makecell{\textbf{52.9} \\ \textcolor{ForestGreen}{\scriptsize $\uparrow$ 18.5}} & \makecell{\textbf{21.2} \\ \textcolor{ForestGreen}{\scriptsize $\uparrow$ 9.1}} & & \makecell{\textbf{70.8} \\ \textcolor{ForestGreen}{\scriptsize $\uparrow$ 14.4}} & \makecell{\textbf{70.7} \\ \textcolor{ForestGreen}{\scriptsize $\uparrow$ 25.0}} & \makecell{\textbf{69.5} \\ \textcolor{ForestGreen}{\scriptsize $\uparrow$ 22.3}} & \makecell{\textbf{26.1} \\ \textcolor{ForestGreen}{\scriptsize $\uparrow$ 10.0}} \\
\bottomrule
\end{tabular}%
}
\caption{Dataset-level macro averaged results.}
\label{tab:dataset_macro_avg}
\end{table*}
\begin{table*}[th]
\centering
\small
\setlength{\tabcolsep}{3pt} 
\resizebox{\textwidth}{!}{%
\begin{tabular}{l c cccc c cccc c cccc c cccc} 
\toprule
\multicolumn{1}{c}{\multirow{2}{*}{\textbf{Models}}} & & \multicolumn{4}{c}{\textbf{SP (\%)~$\uparrow$}} & & \multicolumn{4}{c}{\textbf{FP (\%)~$\uparrow$}} & & \multicolumn{4}{c}{\textbf{SPA (\%)~$\uparrow$}} & & \multicolumn{4}{c}{\textbf{FPA (\%)~$\uparrow$}} \\
\cmidrule(lr){3-6} \cmidrule(lr){8-11} \cmidrule(lr){13-16} \cmidrule(lr){18-21}
 & & SH & MH & ST & MT & & SH & MH & ST & MT & & SH & MH & ST & MT & & SH & MH & ST & MT \\
\midrule
\rowcolor{gray!15} \multicolumn{21}{c}{\textit{\textbf{Hybrid-15 Setting}}} \\
\midrule
Qwen3-8B (Vanilla) & & 71.3 & 24.5 & 67.5 & 0.0 & & 71.3 & 56.6 & 70.8 & 33.3 & & 21.8 & 40.7 & 24.0 & 17.1 & & 39.3 & 49.7 & 40.9 & 21.0 \\
\rowcolor{TrueLightPurple} 
\textbf{UniToolCall (Ours)} & & \makecell{\textbf{95.2} \\ \textcolor{ForestGreen}{\scriptsize $\uparrow$ 23.9}} & \makecell{\textbf{78.6} \\ \textcolor{ForestGreen}{\scriptsize $\uparrow$ 54.1}} & \makecell{\textbf{94.9} \\ \textcolor{ForestGreen}{\scriptsize $\uparrow$ 27.4}} & \makecell{0.0 \\ \textcolor{WildStrawberry}{\scriptsize $\downarrow$ 0.0}} & & \makecell{\textbf{95.2} \\ \textcolor{ForestGreen}{\scriptsize $\uparrow$ 23.9}} & \makecell{\textbf{88.8} \\ \textcolor{ForestGreen}{\scriptsize $\uparrow$ 32.2}} & \makecell{\textbf{95.8} \\ \textcolor{ForestGreen}{\scriptsize $\uparrow$ 25.0}} & \makecell{\textbf{41.8} \\ \textcolor{ForestGreen}{\scriptsize $\uparrow$ 8.5}} & & \makecell{\textbf{28.0} \\ \textcolor{ForestGreen}{\scriptsize $\uparrow$ 6.2}} & \makecell{\textbf{66.4} \\ \textcolor{ForestGreen}{\scriptsize $\uparrow$ 25.7}} & \makecell{\textbf{32.4} \\ \textcolor{ForestGreen}{\scriptsize $\uparrow$ 8.4}} & \makecell{\textbf{18.5} \\ \textcolor{ForestGreen}{\scriptsize $\uparrow$ 1.4}} & & \makecell{\textbf{49.9} \\ \textcolor{ForestGreen}{\scriptsize $\uparrow$ 10.6}} & \makecell{\textbf{77.4} \\ \textcolor{ForestGreen}{\scriptsize $\uparrow$ 27.7}} & \makecell{\textbf{53.4} \\ \textcolor{ForestGreen}{\scriptsize $\uparrow$ 12.5}} & \makecell{\textbf{28.4} \\ \textcolor{ForestGreen}{\scriptsize $\uparrow$ 7.4}} \\
\midrule
\rowcolor{gray!15} \multicolumn{21}{c}{\textit{\textbf{Hybrid-25 Setting}}} \\
\midrule
Qwen3-8B (Vanilla) & & 65.5 & 22.4 & 61.9 & 0.0 & & 65.5 & 51.5 & 64.9 & 29.2 & & 19.5 & 37.3 & 21.6 & 15.3 & & 35.3 & 45.1 & 36.8 & 18.0 \\
\rowcolor{TrueLightPurple} 
\textbf{UniToolCall (Ours)} & & \makecell{\textbf{94.6} \\ \textcolor{ForestGreen}{\scriptsize $\uparrow$ 29.1}} & \makecell{\textbf{81.2} \\ \textcolor{ForestGreen}{\scriptsize $\uparrow$ 58.8}} & \makecell{\textbf{94.6} \\ \textcolor{ForestGreen}{\scriptsize $\uparrow$ 32.7}} & \makecell{0.0 \\ \textcolor{WildStrawberry}{\scriptsize $\downarrow$ 0.0}} & & \makecell{\textbf{94.6} \\ \textcolor{ForestGreen}{\scriptsize $\uparrow$ 29.1}} & \makecell{\textbf{90.4} \\ \textcolor{ForestGreen}{\scriptsize $\uparrow$ 38.9}} & \makecell{\textbf{95.4} \\ \textcolor{ForestGreen}{\scriptsize $\uparrow$ 30.5}} & \makecell{\textbf{37.8} \\ \textcolor{ForestGreen}{\scriptsize $\uparrow$ 8.6}} & & \makecell{\textbf{27.9} \\ \textcolor{ForestGreen}{\scriptsize $\uparrow$ 8.4}} & \makecell{\textbf{67.1} \\ \textcolor{ForestGreen}{\scriptsize $\uparrow$ 29.8}} & \makecell{\textbf{32.4} \\ \textcolor{ForestGreen}{\scriptsize $\uparrow$ 10.8}} & \makecell{\textbf{16.0} \\ \textcolor{ForestGreen}{\scriptsize $\uparrow$ 0.7}} & & \makecell{\textbf{49.5} \\ \textcolor{ForestGreen}{\scriptsize $\uparrow$ 14.2}} & \makecell{\textbf{78.7} \\ \textcolor{ForestGreen}{\scriptsize $\uparrow$ 33.6}} & \makecell{\textbf{53.2} \\ \textcolor{ForestGreen}{\scriptsize $\uparrow$ 16.4}} & \makecell{\textbf{24.4} \\ \textcolor{ForestGreen}{\scriptsize $\uparrow$ 6.4}} \\
\bottomrule
\end{tabular}%
}
\caption{Ablation on candidate pool size. Evaluation results under Hybrid-15 and Hybrid-25 settings.}
\label{tab:hybrid_15_25}
\end{table*}
\begin{table*}[t]
\centering
\small
\setlength{\tabcolsep}{3pt} 
\resizebox{\textwidth}{!}{%
\begin{tabular}{l c cccc c cccc c cccc c cccc} 
\toprule
\multicolumn{1}{c}{\multirow{2}{*}{\textbf{Proxy-Scale Models (20K)}}} & & \multicolumn{4}{c}{\textbf{SP (\%)~$\uparrow$}} & & \multicolumn{4}{c}{\textbf{FP (\%)~$\uparrow$}} & & \multicolumn{4}{c}{\textbf{SPA (\%)~$\uparrow$}} & & \multicolumn{4}{c}{\textbf{FPA (\%)~$\uparrow$}} \\
\cmidrule(lr){3-6} \cmidrule(lr){8-11} \cmidrule(lr){13-16} \cmidrule(lr){18-21}
 & & SH & MH & ST & MT & & SH & MH & ST & MT & & SH & MH & ST & MT & & SH & MH & ST & MT \\
\midrule
\rowcolor{gray!15} \multicolumn{21}{c}{\textit{\textbf{Ablation Group 1: Impact of Toucan Dominance \& Synthetic Data}}} \\
\midrule
Toucan-Only Random & & 87.8 & 16.5 & 81.7 & 2.8 & & 87.8 & 55.7 & 85.8 & 37.6 & & 26.3 & 41.2 & 28.3 & 15.8 & & 47.8 & 50.9 & 48.8 & 21.5 \\
Pure Public Random & & \underline{89.9} & \underline{22.1} & 84.1 & \textbf{5.6} & & \underline{89.9} & 58.1 & 87.8 & \textbf{40.9} & & 26.6 & 44.7 & 28.9 & 17.0 & & 48.8 & 53.4 & \underline{50.0} & \textbf{24.6} \\
Syn + Pub Random & & 90.0 & \textbf{22.4} & \underline{84.2} & \textbf{5.6} & & 90.0 & \textbf{59.3} & \underline{88.1} & 39.5 & & \underline{27.0} & \textbf{45.6} & \underline{29.3} & \textbf{18.7} & & \underline{48.9} & \textbf{54.6} & 50.2 & \underline{22.8} \\
Syn + Pub Balanced & & \textbf{93.4} & 21.0 & \textbf{87.1} & \textbf{5.6} & & \textbf{93.4} & \underline{58.4} & \textbf{91.0} & \underline{40.6} & & \textbf{27.9} & \textbf{45.6} & \textbf{30.1} & \underline{18.2} & & \textbf{50.7} & \underline{54.0} & \textbf{51.8} & 23.4 \\
\midrule
\rowcolor{gray!15} \multicolumn{21}{c}{\textit{\textbf{Ablation Group 2: Zero-Shot Generalization to Unseen Tools}}} \\
\midrule
Strictly Unseen Tools & & 89.8 & 22.4 & 84.1 & 5.6 & & 89.8 & 60.4 & 88.0 & 41.6 & & 27.2 & 47.1 & 29.7 & 17.2 & & 49.2 & 55.8 & 50.6 & 25.4 \\
\bottomrule
\end{tabular}%
}
\caption{Proxy-scale controlled ablations under a matched 20K data budget.}
\label{tab:proxy_ablations}
\end{table*}
\begin{figure*}[t]
    \centering
    \includegraphics[width=\textwidth]{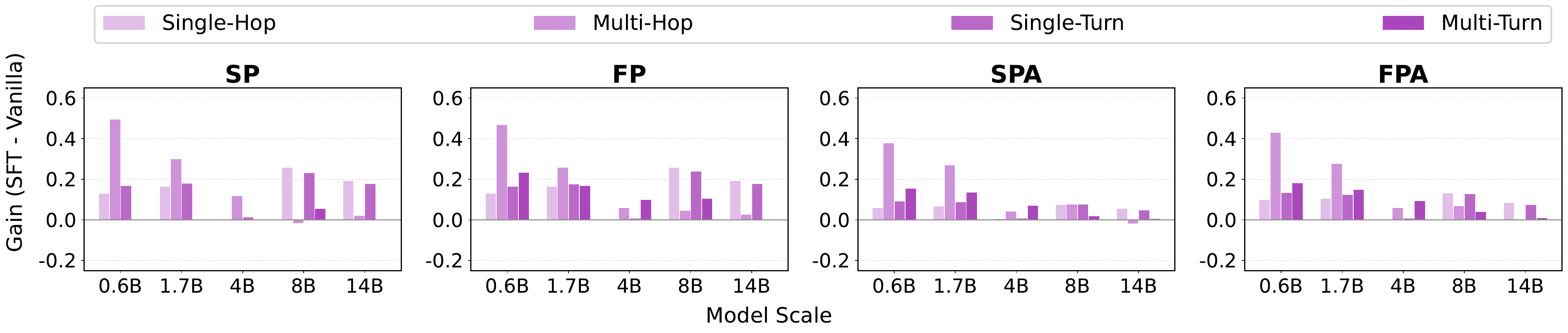}
    \caption{Performance gains of our SFT pipeline over the vanilla backbones across Qwen3 scales (0.6B-14B) under the Syn+Pub Balanced 20K recipe.}
    \label{fig:scaling_laws}
\end{figure*}

\begin{table*}[t]
\centering
\footnotesize
\setlength{\tabcolsep}{3pt}
\resizebox{\textwidth}{!}{%
\begin{tabular}{l c cccc c cccc c cccc c cccc}
\toprule
\multicolumn{1}{c}{\multirow{2}{*}{\textbf{Model}}} & & \multicolumn{4}{c}{\textbf{SP (\%)~$\uparrow$}} & & \multicolumn{4}{c}{\textbf{FP (\%)~$\uparrow$}} & & \multicolumn{4}{c}{\textbf{SPA (\%)~$\uparrow$}} & & \multicolumn{4}{c}{\textbf{FPA (\%)~$\uparrow$}} \\
\cmidrule(lr){3-6} \cmidrule(lr){8-11} \cmidrule(lr){13-16} \cmidrule(lr){18-21}
 & & SH & MH & ST & MT & & SH & MH & ST & MT & & SH & MH & ST & MT & & SH & MH & ST & MT \\
\midrule
\rowcolor{gray!15} \multicolumn{21}{c}{\textit{\textbf{Qwen3-0.6B}}} \\
\midrule
Vanilla & & 56.3 & 0.8 & 51.4 & 0.0 & & 56.3 & 26.1 & 54.1 & 14.6 & & 15.2 & 17.2 & 15.7 & 5.0 & & 28.2 & 22.8 & 28.1 & 8.6 \\
\rowcolor{TrueLightPurple}
SFT & & 69.4 & \underline{50.3} & 68.3 & 0.0 & & 69.4 & 72.9 & 70.6 & 38.1 & & 21.2 & 55.0 & 25.0 & 20.6 & & 38.3 & 65.9 & 41.6 & 26.8 \\
\midrule
\rowcolor{gray!15} \multicolumn{21}{c}{\textit{\textbf{Qwen3-1.7B}}} \\
\midrule
Vanilla & & 52.3 & 19.6 & 49.8 & 0.0 & & 52.3 & 47.4 & 52.6 & 21.2 & & 14.5 & 28.2 & 16.2 & 7.2 & & 27.5 & 38.3 & 29.1 & 11.9 \\
\rowcolor{TrueLightPurple}
SFT & & 68.9 & 49.7 & 67.8 & 0.0 & & 68.9 & 73.2 & 70.2 & 38.1 & & 21.3 & \underline{55.3} & 25.1 & 20.8 & & 38.2 & 66.1 & 41.5 & 26.8 \\
\midrule
\rowcolor{gray!15} \multicolumn{21}{c}{\textit{\textbf{Qwen3-4B}}} \\
\midrule
Vanilla & & 68.5 & 38.4 & 66.4 & 0.0 & & 68.5 & 67.5 & 69.4 & 27.5 & & 20.6 & 50.8 & 24.1 & 13.9 & & 37.2 & 60.5 & 40.3 & 17.8 \\
\rowcolor{TrueLightPurple}
SFT & & 68.8 & \underline{50.3} & 67.8 & 0.0 & & 68.8 & \underline{73.5} & 70.2 & 37.5 & & 21.1 & 55.0 & 24.9 & \underline{20.9} & & 37.7 & \underline{66.5} & 41.1 & \underline{27.1} \\
\midrule
\rowcolor{gray!15} \multicolumn{21}{c}{\textit{\textbf{Qwen3-8B}}} \\
\midrule
Vanilla & & 67.6 & 22.8 & 63.9 & 0.0 & & 67.6 & 53.8 & 67.1 & 30.0 & & 20.3 & 37.9 & 22.4 & 16.2 & & 37.4 & 47.1 & 38.9 & 19.3 \\
\rowcolor{TrueLightPurple}
SFT & & \textbf{93.4} & 21.0 & \underline{87.1} & \textbf{5.6} & & \textbf{93.4} & 58.4 & \underline{91.0} & \underline{40.6} & & \underline{27.9} & 45.6 & \underline{30.1} & 18.2 & & \textbf{50.7} & 54.0 & \underline{51.8} & 23.4 \\
\midrule
\rowcolor{gray!15} \multicolumn{21}{c}{\textit{\textbf{Qwen3-14B}}} \\
\midrule
Vanilla & & 73.9 & 65.6 & 74.1 & 0.0 & & 73.9 & 81.6 & 75.6 & 43.0 & & 23.2 & 62.5 & 27.6 & 24.2 & & 42.1 & 73.6 & 45.9 & 29.4 \\
\rowcolor{TrueLightPurple}
SFT & & \underline{93.3} & \textbf{67.8} & \textbf{91.9} & 0.0 & & \underline{93.3} & \textbf{84.3} & \textbf{93.6} & \textbf{43.0} & & \textbf{28.8} & \textbf{60.5} & \textbf{32.4} & \textbf{24.9} & & \underline{50.5} & \textbf{73.1} & \textbf{53.5} & \textbf{30.4} \\
\bottomrule
\end{tabular}%
}
\caption{Scaling across Qwen3 model sizes. Hybrid-20 scores under the Syn+Pub Balanced 20K recipe.}
\label{tab:qwen_scale_raw}
\end{table*}

\begin{table*}[t]
\centering
\footnotesize
\setlength{\tabcolsep}{3pt}
\resizebox{\textwidth}{!}{%
\begin{tabular}{l c cccc c cccc c cccc c cccc}
\toprule
\multicolumn{1}{c}{\multirow{2}{*}{\textbf{Model}}} & & \multicolumn{4}{c}{\textbf{SP (\%)~$\uparrow$}} & & \multicolumn{4}{c}{\textbf{FP (\%)~$\uparrow$}} & & \multicolumn{4}{c}{\textbf{SPA (\%)~$\uparrow$}} & & \multicolumn{4}{c}{\textbf{FPA (\%)~$\uparrow$}} \\
\cmidrule(lr){3-6} \cmidrule(lr){8-11} \cmidrule(lr){13-16} \cmidrule(lr){18-21}
 & & SH & MH & ST & MT & & SH & MH & ST & MT & & SH & MH & ST & MT & & SH & MH & ST & MT \\
\midrule
\rowcolor{gray!15} \multicolumn{21}{c}{\textit{\textbf{Gain $=$ SFT $-$ Vanilla}}} \\
\midrule
0.6B & & 13.1 & \textbf{49.5} & 16.9 & 0.0 & & 13.1 & \textbf{46.8} & 16.5 & \textbf{23.4} & & 6.0 & \textbf{37.9} & \textbf{9.3} & \textbf{15.6} & & 10.1 & \textbf{43.1} & \textbf{13.4} & \textbf{18.2} \\
1.7B & & 16.5 & \underline{30.1} & \underline{18.1} & 0.0 & & 16.5 & \underline{25.8} & 17.5 & \underline{16.9} & & \underline{6.8} & \underline{27.1} & \underline{8.9} & \underline{13.5} & & \underline{10.6} & \underline{27.8} & 12.4 & \underline{14.9} \\
4B & & 0.4 & 11.9 & 1.4 & 0.0 & & 0.4 & 6.0 & 0.8 & 10.0 & & 0.5 & 4.2 & 0.8 & 7.0 & & 0.5 & 5.9 & 0.8 & 9.3 \\
8B & & \textbf{25.8} & $-$1.8 & \textbf{23.2} & \textbf{5.6} & & \textbf{25.8} & 4.6 & \textbf{23.9} & 10.6 & & \textbf{7.5} & 7.7 & 7.7 & 2.0 & & \textbf{13.3} & 6.9 & \underline{12.9} & 4.1 \\
14B & & \underline{19.4} & 2.2 & 17.8 & 0.0 & & \underline{19.4} & 2.7 & \underline{18.0} & 0.0 & & 5.6 & $-$2.0 & 4.8 & 0.7 & & 8.4 & $-$0.5 & 7.6 & 1.0 \\
\bottomrule
\end{tabular}%
}
\caption{Absolute gains of SFT over Vanilla across Qwen3 scales.}
\label{tab:qwen_scale_gain}
\end{table*}

\begin{table*}[t]
\centering
\footnotesize
\setlength{\tabcolsep}{3pt}
\resizebox{\textwidth}{!}{%
\begin{tabular}{l c cccc c cccc c cccc c cccc}
\toprule
\multicolumn{1}{c}{\multirow{2}{*}{\textbf{Model}}} & & \multicolumn{4}{c}{\textbf{SP (\%)~$\uparrow$}} & & \multicolumn{4}{c}{\textbf{FP (\%)~$\uparrow$}} & & \multicolumn{4}{c}{\textbf{SPA (\%)~$\uparrow$}} & & \multicolumn{4}{c}{\textbf{FPA (\%)~$\uparrow$}} \\
\cmidrule(lr){3-6} \cmidrule(lr){8-11} \cmidrule(lr){13-16} \cmidrule(lr){18-21}
 & & SH & MH & ST & MT & & SH & MH & ST & MT & & SH & MH & ST & MT & & SH & MH & ST & MT \\
\midrule
\rowcolor{gray!15} \multicolumn{21}{c}{\textit{\textbf{Gemma-3-4B}}} \\
\midrule
Vanilla & & 62.3 & 9.1 & 57.7 & 0.0 & & 62.3 & 32.4 & 60.2 & 12.7 & & 16.0 & 17.3 & 16.4 & 4.6 & & 29.8 & 25.5 & 29.8 & 6.9 \\
\rowcolor{TrueLightPurple}
SFT & & \textbf{95.5} & \textbf{20.7} & \textbf{89.1} & 0.0 & & \textbf{95.5} & \textbf{52.8} & \textbf{92.5} & \textbf{29.8} & & \textbf{27.4} & \textbf{28.7} & \textbf{28.0} & \textbf{12.8} & & \textbf{49.4} & \textbf{39.0} & \textbf{49.2} & \textbf{14.0} \\
\midrule
\rowcolor{gray!15} \multicolumn{21}{c}{\textit{\textbf{Llama-3.2-3B}}} \\
\midrule
Vanilla & & 55.8 & 3.1 & 51.1 & 0.0 & & 55.8 & 31.2 & 54.0 & 27.7 & & 11.4 & 17.5 & 12.2 & 10.0 & & 24.8 & 24.0 & 25.1 & 14.9 \\
\rowcolor{TrueLightPurple}
SFT & & \textbf{89.4} & \textbf{21.0} & \textbf{83.6} & 0.0 & & \textbf{89.4} & \textbf{53.1} & \textbf{86.9} & \textbf{43.4} & & \textbf{25.5} & \textbf{30.3} & \textbf{26.3} & \textbf{19.3} & & \textbf{45.7} & \textbf{40.6} & \textbf{45.8} & \textbf{26.5} \\
\bottomrule
\end{tabular}%
}
\caption{Cross-family evaluation on Gemma-3-4B and Llama-3.2-3B. Hybrid-20 scores under the Syn+Pub Balanced 20K recipe.}
\label{tab:cross_family_raw}
\end{table*}

\begin{table*}[t]
\centering
\footnotesize
\setlength{\tabcolsep}{3pt}
\resizebox{\textwidth}{!}{%
\begin{tabular}{l c cccc c cccc c cccc c cccc}
\toprule
\multicolumn{1}{c}{\multirow{2}{*}{\textbf{Model}}} & & \multicolumn{4}{c}{\textbf{SP (\%)~$\uparrow$}} & & \multicolumn{4}{c}{\textbf{FP (\%)~$\uparrow$}} & & \multicolumn{4}{c}{\textbf{SPA (\%)~$\uparrow$}} & & \multicolumn{4}{c}{\textbf{FPA (\%)~$\uparrow$}} \\
\cmidrule(lr){3-6} \cmidrule(lr){8-11} \cmidrule(lr){13-16} \cmidrule(lr){18-21}
 & & SH & MH & ST & MT & & SH & MH & ST & MT & & SH & MH & ST & MT & & SH & MH & ST & MT \\
\midrule
\rowcolor{gray!15} \multicolumn{21}{c}{\textit{\textbf{Gain $=$ SFT $-$ Vanilla}}} \\
\midrule
Gemma-3-4B & & 33.2 & 11.6 & 31.4 & 0.0 & & 33.2 & 20.4 & 32.3 & \textbf{17.1} & & 11.4 & 11.4 & 11.6 & 8.2 & & 19.6 & 13.5 & 19.4 & 7.1 \\
Llama-3.2-3B & & \textbf{33.6} & \textbf{17.9} & \textbf{32.5} & 0.0 & & \textbf{33.6} & \textbf{21.9} & \textbf{32.9} & 15.7 & & \textbf{14.1} & \textbf{12.8} & \textbf{14.1} & \textbf{9.3} & & \textbf{20.9} & \textbf{16.6} & \textbf{20.7} & \textbf{11.6} \\
\bottomrule
\end{tabular}%
}
\caption{Absolute gains of SFT over Vanilla for non-Qwen backbones.}
\label{tab:cross_family_gain}
\end{table*}

\subsection{Run-level stability statistics}
\label{sec:appendix_stability}

Table~\ref{tab:repeated_run_stability} reports repeated-run statistics for the trainable backbone models. Both \method and the vanilla Qwen3-8B are trained three times with independent runs. The reported values are mean $\pm$ sample standard deviation across runs. The gains of \method over Vanilla hold across runs, particularly for single/multi-hop and single-turn tool selection, where the variance of \method is also smaller. Multi-turn metrics exhibit larger fluctuations due to the small evaluation size and the inherent difficulty of strict conversation-level matching.

\subsection{Dataset-level macro-average analysis}
\label{sec:appendix_imbalance}
To mitigate the sample imbalance in $D_{\text{test}}$ and assess true cross-domain generalization, we compute a dataset-level macro-average. This unweighted mean across the 7 sub-benchmarks grants equal weight to each dataset, explicitly neutralizing HammerBench's dominance.

As shown in Table~\ref{tab:dataset_macro_avg}, \method\ achieves 81.0\% Single-Turn SP and 72.8\% Multi-Hop SP, significantly outperforming the vanilla baseline's 45.8\% and 31.2\%, respectively. This pattern across equally weighted evaluation distributions indicates that the gains are not confined to the largest source dataset.

\subsection{Robustness to Candidate Pool Size}
\label{sec:hybrid_15_25}
To investigate whether the model overfits to a fixed 20-tool setting and to evaluate its generalization against varying distractor volumes, we conduct supplementary evaluations under Hybrid-15 and Hybrid-25 configurations. As shown in Table~\ref{tab:hybrid_15_25}, \method remains accurate when the pool shrinks to 15 or expands to 25, and the gap over vanilla Qwen3-8B holds across metrics. The gains therefore do not appear to depend on a fixed candidate-list length.

\subsection{Disentangling Data Components and Unseen Generalization}
\label{sec:unseen_generalization}
To isolate the impact of data composition and zero-shot generalization, we conduct proxy-scale ablations under a matched 20K budget (Table~\ref{tab:proxy_ablations}). First, Syn+Pub Balanced significantly outperforms Toucan-Only (e.g., 21.0\% vs. 16.5\% Multi-hop SP), indicating that structural diversity drives performance more than sheer scale. Second, injecting a 2.9K synthetic subset (Syn+Pub Random) into the public baseline enhances parameter grounding (+0.9\% Multi-hop SPA), suggesting synthetic trajectories act as an effective structural catalyst. Finally, to address schema leakage, we evaluate a Strictly Unseen Tools model that removes test APIs from training candidates. Its performance matches the exposed model (22.4\% Multi-hop SP in both), demonstrating robust zero-shot generalization to novel APIs.

\subsection{Scaling Across Model Sizes and Families}
\label{sec:scaling_laws}
\label{sec:cross_family}
To assess whether the pipeline remains effective beyond the primary 8B backbone, we evaluate the Syn+Pub Balanced 20K recipe across Qwen3 from 0.6B to 14B under Hybrid-20 (Figure~\ref{fig:scaling_laws}; Tables~\ref{tab:qwen_scale_raw} and~\ref{tab:qwen_scale_gain}). Single-hop and single-turn metrics improve at every scale. Capacity still matters: 0.6B/1.7B gain most on multi-hop because Vanilla is near floor; 4B is already strong on single-hop SP (68.5\%, +0.4\,pp); 8B has the largest Strict Precision gains (+25.8 / +23.2\,pp) while multi-hop SP does not rise (21.0\% vs.\ 22.8\% Vanilla); 14B continues to improve on single-hop / single-turn (+19.4 / +17.8\,pp). Unevenness is largely confined to multi-hop. Due to compute constraints, 14B is our current fine-tuning upper bound.

The same recipe on Gemma-3-4B-it and Llama-3.2-3B-Instruct yields +33.2 / +31.4\,pp and +33.6 / +32.5\,pp on single-hop / single-turn SP (Tables~\ref{tab:cross_family_raw} and~\ref{tab:cross_family_gain}).

\subsection{Multi-turn strict-precision cliff analysis}
\label{sec:appendix_mt_cliff}

Conversation-level Multi-Turn Strict Precision (MT SP) is an all-or-nothing metric: a single incorrect function call zeros the entire dialogue. Under Hybrid-20, both Vanilla and \method therefore score 0.0\% MT SP on the full test suite, even though Flexible multi-turn metrics remain non-trivial. This pattern reflects error accumulation under strict episode scoring rather than an inability to select correct tools on individual turns.

To diagnose the cliff, we analyze 30 conversations from the BFCL v3 multi-turn subset. We report Prefix Survival: the fraction of dialogues that remain fully correct on Turns $1,\ldots,k$. Once any earlier turn fails, the dialogue contributes 0 to all longer prefixes. Shorter dialogues drop out of later-$k$ estimates, which is why the annotated $n$ decreases with $k$.

\begin{figure}[t]
    \centering
    \includegraphics[width=\columnwidth]{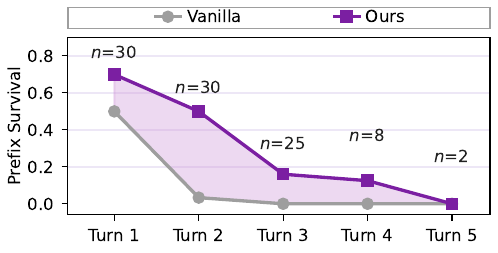}
    \caption{Prefix Survival on the BFCL v3 multi-turn subset under Hybrid-20.}
    \label{fig:mt_turn_cliff}
\end{figure}

Figure~\ref{fig:mt_turn_cliff} visualizes this accumulation. Vanilla collapses by Turn 2 (3.3\%), whereas \method remains at 50.0\%. Later-turn estimates become noisy as the eligible conversation count shrinks; we therefore emphasize the Turn-2 gap rather than the small-$n$ tail.
\begin{table}[t]
\centering
\small
\setlength{\tabcolsep}{6pt}
\begin{tabular*}{\columnwidth}{@{\extracolsep{\fill}}lrrr@{}}
\toprule
\textbf{Severity} & \textbf{Vanilla (\%)} & \textbf{Ours (\%)} & \textbf{$\Delta$ (pp)} \\
\midrule
Catastrophic & 30.0 & 0.0 & $-$30.0 \\
Hard & 43.3 & 53.3 & +10.0 \\
Soft / near-miss & 26.7 & 46.7 & +20.0 \\
\bottomrule
\end{tabular*}
\caption{First-error severity on 30 BFCL multi-turn conversations.}
\label{tab:mt_first_error}
\end{table}

On the same 30 conversations, we inspect first-error type using Qwen3-32B classification with human verification (Table~\ref{tab:mt_first_error}). Rows correspond to format/parse failure (Catastrophic), wrong tool selection (Hard), and wrong arguments or arity (Soft / near-miss). \method eliminates catastrophic failures (30.0\% $\rightarrow$ 0.0\%); remaining errors shift toward near-misses. The first strict error also occurs later on average (Turn 1.40 vs.\ 1.80). Long-horizon strict success nonetheless remains an open challenge; as a data-side mitigation, our synthetic pipeline enforces cross-turn dependencies via Anchor Linkage (Appendix~\ref{sec:appendix_anchor_linkage}).

\subsection{Example of the action-only training format}
\label{sec:data_sample}

Observation and Answer fields are retained in the dataset for evaluation purposes but are not part of the prediction target during fine-tuning. This design isolates the model's tool-selection and parameter-generation capabilities from downstream response realization. A full single-hop sample is shown on the following page.

\clearpage
\onecolumn

\definecolor{DeepPurple}{HTML}{5E2A84}   
\definecolor{LightPurple}{HTML}{F9F4FA}  

\begin{tcblisting}{
    colback=LightPurple,  
    colframe=DeepPurple, 
    title=Data Sample ($P_{\text{sys}}$), 
    arc=4pt, 
    boxrule=1pt,
    fonttitle=\bfseries, 
    breakable,
    listing only, 
    width=\textwidth,            
    listing options={
        basicstyle=\ttfamily\scriptsize, 
        breaklines=true,                 
        breakatwhitespace=false, 
        columns=fullflexible,            
        keepspaces=true,                 
        showstringspaces=false           
    }
}
{
  "conversations":[
    {
      "from": "human",
      "value": "Provide secure access to medical records for a patient named John Smith."
    },
    {
      "from": "function_call",
      "value": "{\"name\":\"MedicalRecordAccess\",\"arguments\":{\"patient_name\":\"John Smith\"}}"
    },
    {
      "from": "observation",
      "value": ""
    },
    {
      "from": "gpt",
      "value": "<answer></answer>"
    }
  ],
  "system": "# Role\n\nYou are an AI assistant capable of calling various functions to help users solve their problems.\n\n# Tool Selection\n\n
  **Important**: The available function signatures are provided in the <tools></tools> section. You must carefully select one or more appropriate tools from this section that can solve the user's request.\n\n
  # Output Rules\n\nYou must strictly follow the rules below when responding:\n\n
  ## 1. Function Call Format\nWhen you need to call a function, you must output only one function call per round in the following format:\n<tool_call>\n{\"name\": <function-name>, \"arguments\": <args-json-object>}\n</tool_call>\n\n
  **Parameter Parsing**: The arguments must be parsed based on the user's query. **Do not fabricate parameters that are not mentioned or cannot be reasonably inferred from the query.** Only use parameters that can be reasonably extracted or inferred from the user's request.\n\n
  **Basis for Generating Function Call Content**:\n- **First function call**: The user's query and available tools information.\n- **n-th function call (n > 1)**: The user's query, available tools information, and the complete conversation history in <chat_history></chat_history> from the previous n-1 rounds (including all prior function calls, observations, and answers). In some scenarios, observations may be empty; this is acceptable for generating function calls.\n\n
  **Example**:\n<tool_call>\n{\\\"name\\\": \\\"cancel_booking\\\", \\\"arguments\\\": {\\\"access_token\\\": \\\"abc123xyz\\\", \\\"booking_id\\\": \\\"flight_001\\\"}}\n</tool_call>\n\n
  ## 2. Answer Format\nWhen you judge from the chat history that all necessary tools have been called, you must immediately stop calling tools and provide the final answer in the following format:\n<answer>\nYour final answer here\n</answer>\n\n
  **Answer Generation Requirements**:\n- **Critical**: If all observations in chat-history are empty (meaning tools were called but returned no data), you MUST reply exactly: \"Sorry, I did not obtain sufficient information to complete your request.\" Do NOT fabricate, invent, or generate any content based on assumptions. Do NOT create imaginary results or responses. Only output this exact message.\n- **Important**: The provided tools may include tools that are irrelevant or unsuitable for the current query. If you determine there are no suitable tools to answer the user's request, reply: \"Sorry, there are no suitable tools to answer your request.\"\n- **Important**: If you have called some tools and obtained observations, but the available tools are insufficient to fully satisfy the user's request (e.g., some required tools are missing from the available tool list), you MUST reply exactly: \"Sorry, there are not enough tools to fully satisfy your request.\" Do NOT fabricate or generate partial answers based on incomplete information.\n- Carefully analyze the conversation history to determine the current turn. The answer must be based on the user's query and all available observation results in the conversation.\n\n
  ## 3. Intelligent Process Stage Judgment\n- single-hop: Typically requires only one tool call to complete the task.\n- multi-hop: Requires multiple tool calls to complete the task.\n- single-turn: Involves only one user query.\n- multi-turn: Involves multiple user queries; later queries may refer to or build upon earlier exchanges.\n- When you see that the assistant has issued a tool call and received an observation, that tool call is considered complete.\n\n
  **Special Note**: By examining the conversation history, you can clearly see:\n- Previous interactions between the user and the assistant\n- Tool calls that have already been executed\n- Results returned by tools\n- The stage the current conversation has reached\n\n
  ## 4. Strictly Prohibited Behaviors\n- Do not output a function call and an answer in the same round.\n- Do not repeatedly call the same tool with identical parameters.\n- Do not ignore existing tool calls and their returned information in the conversation history.\n- Do not fabricate parameters that are not present in or reasonably implied by the user's query.\n\n
  ## 5. Error Handling and Data Quality Assessment\n- If the tool returns an empty observation, it may indicate there is no data under the current query conditions or that observation data is unavailable in the current context.\n- If the tool returns error messages (e.g., \"resource not found\", \"invalid parameters\"), do not repeat the same tool call.\n- In such cases, provide an explanatory answer describing the specific error cause or data condition.\n- Absolutely do not repeatedly call the same tool because it returned an error or empty data.\n\n\n",
  "tools": "[\n    {\"name\": \"MedicalRecordAccess\", \n     \"description\": \"API for providing secure access to medical records.\", \n     \"category\": \"operations\", \n     \"domain\": \"healthcare\", \n     \"inputSchema\": {\"type\": \"object\", \"properties\": {\"patient_name\": {\"type\": \"str\", \"description\": \"The name of the patient.\"}}, \"required\": [\"patient_name\"]}\n    }, \n  ... [18 distractor tools omitted for brevity] ..., \n    \n    {\"name\": \"update_with_defaults\", \n     \"description\": \"Updates the defaults dictionary with the values from the updates dictionary.\", \n     \"inputSchema\": {\"type\": \"object\", \"properties\": {\"defaults\": {\"type\": \"object\", \"additionalProperties\": {\"type\": \"integer\"}, \"description\": \"The default dictionary to be updated.\"}, \"updates\": {\"type\": \"object\", \"additionalProperties\": {\"type\": \"integer\"}, \"description\": \"The dictionary containing updates to apply to the defaults.\"}}, \"required\": [\"defaults\", \"updates\"]}, \n     \"category\": \"operations\", \n     \"domain\": \"technology\"\n    }\n]"
}
\end{tcblisting}

\end{document}